\documentclass{article}

\usepackage[preprint]{neurips_2023} 

\usepackage[resetlabels]{multibib}

\usepackage[utf8]{inputenc} 
\usepackage[T1]{fontenc}    
\usepackage{hyperref}       
\usepackage{url}            
\usepackage{booktabs}       
\usepackage{amsfonts}       
\usepackage{nicefrac}       
\usepackage{microtype}      
\usepackage{xcolor}         
\usepackage{tablefootnote}

\usepackage{placeins}
\usepackage{amssymb}
\usepackage{amsmath,amsfonts,amssymb,amscd,amsthm,xspace}
\usepackage{tikz}
\usetikzlibrary{arrows}
\usetikzlibrary{backgrounds}
\usepackage{color}
\usepackage{graphicx}
\usepackage{caption}
\usepackage{subcaption}
\usepackage{float}

\usepackage{array}
\usepackage{hyperref}
\usepackage{longtable}

\usepackage{algorithm}
\usepackage{algpseudocode}

\newcommand\lb{\left}
\newcommand\rb{\right}

\title{Cold PAWS: Unsupervised class discovery and addressing the cold-start problem for semi-supervised learning}

\author{%
  Evelyn J. Mannix \\
  Melbourne Centre for Data Science\\
  Univeristy of Melbourne\\
  Melbourne, Australia 3010 \\
  \texttt{evelyn.mannix@unimelb.edu.au} \\
  \And
  Howard Bondell \\
  Melbourne Centre for Data Science\\
  Univeristy of Melbourne\\
  Melbourne, Australia 3010 \\
  \texttt{howard.bondell@unimelb.edu.au} \\
}

\begin{document}

\maketitle

\begin{abstract}
In many machine learning applications, labeling datasets can be an arduous and time-consuming task. Although research has shown that semi-supervised learning techniques can achieve high accuracy with very few labels within the field of computer vision, little attention has been given to how images within a dataset should be selected for labeling. In this paper, we propose a novel approach based on well-established self-supervised learning, clustering, and manifold learning techniques that address this challenge of selecting an informative image subset to label in the first instance, which is known as the \textit{cold-start} or \textit{unsupervised selective labelling} problem. We test our approach using several publicly available datasets, namely CIFAR10, Imagenette, DeepWeeds, and EuroSAT, and observe improved performance with both supervised and semi-supervised learning strategies when our label selection strategy is used, in comparison to random sampling. We also obtain superior performance for the datasets considered with a much simpler approach compared to other methods in the literature.
\end{abstract}

\section{Introduction}

Supervised deep learning models benefit from being trained on large amounts of labelled data \citep{kolesnikov2020big}. However, labelling large datasets with sufficient quality is a significant challenge in many computer vision projects. In fields like medical pathology \citep{nguyen2020vindr, bustos2020padchest} and ecology \citep{schneider2020three}, experts may be drawn from a limited pool, or physical tests may be necessary to label images, which presents further challenges. This need for labelled data restricts the use of computer vision technologies, particularly for researchers and others without the resources to label large datasets.

New semi-supervised learning approaches are achieving very promising results on computer vision benchmarking problems, and can achieve results comparable to supervised learning methods in some cases. These breakthroughs include semi-supervised pseudo-labelling methods \citep{sohn2020fixmatch} and self-supervised contrastive learning approaches \citep{chen2020big}, which continue to be surpassed by more recent techniques \citep{wang2022freematch, assran2021semi}.

Although these methods can substantially reduce the number of labels required to build accurate models for particular tasks \citep{sajun2022investigating, taleb2022self}, the decision of which images to label first is left to the practitioner. This is known as the cold-start learning problem, and it has been observed that using prototypical class examples from a dataset can lead to large boosts in semi-supervised performance \citep{wang2022freematch}. However, identifying these images \textit{a priori} remains relatively unexplored in the computer vision literature.

Previous studies have generally considered the cold-start problem in the context of active learning, using self-supervised methods and various other strategies to identify smart initial pools \citep{jin2022cold, yi2022using, ju2022extending, chandra2021initial}. However, the results of these studies are not competitive compared to state-of-the-art semi-supervised learning methods on benchmark problems. It has also been observed that the improvements gained through using active learning methods can be marginal in comparison to self-supervised and semi-supervised approaches \citep{boushehri2020annotation, chan2021marginal}. 

More recent work has referred to this problem as unsupervised selective labelling \citep{wang2022unsupervised}, and demonstrates that selecting particular instances to label using self-supervised embeddings is an effective strategy to boost semi-supervised learning performance. Nevertheless, the strategy proposed within this work adds additional training steps \textemdash{} to obtain the best results it is required to fine-tune the self-supervised model using a clustering inspired loss function.

In this paper, we propose a much simpler approach for addressing the cold-start learning problem. As in previous studies, we apply self-supervised learning approaches to map our image dataset onto a low dimensional feature space, but then use well-established and readily available clustering and manifold learning techniques to select representative images. We are able to show that our approach succeeds in selecting informative images that, compared to random sampling, (i) cover the set of classes of interest in the dataset with fewer samples, and (ii) result in better performing computer vision models, with both supervised and semi-supervised learning techniques, through better sampling of the diversity within each class. We also obtain superior performance compared to previous work.

\section{Background}

\subsection{Self-supervised learning}

Self-supervised learning approaches are a class of methods developed within the field of representation learning. The goal is to learn a representation of a dataset that maximizes performance on unknown downstream tasks. In the computer vision space, there are several approaches to achieve this, including training neural networks with dummy tasks \citep{komodakis2018unsupervised}, bootstrapping \citep{grill2020bootstrap, caron2021emerging}, and contrastive learning \citep{chen2020simple}.

In this work, we focus on one of the breakthrough contrastive learning methods; SimCLR (a simple framework for contrastive learning of visual representations) \citep{chen2020simple}. This method takes a batch of images $x_i$ of size $B$, applies two stochastic transformations to each image to create $2B$ augmented images $x'_i$, and uses a neural network to obtain representations $z_i = f(x'_i)$ in a lower dimensional space. The key idea is to create a loss that teaches the network to be able to match each augmented view in the batch with its positive pair \textemdash{} the one transformed from the same image. For an $i,j$ positive pair, this loss is given as:

\begin{align}
    L_{ij} = -\log  \frac{e^{\delta(z_i, z_j)/T}}{\sum_{k=1, k\not=i}^{2N} e^{\delta(z_i, z_k)/T}}
    \label{eq:NT-Xent}
\end{align}

where the function $\delta$ is defined by the cosine similarity $\delta(z_i, z_j) = \frac{z_i.z_j}{||z_i||||z_j||}$ and $T$ is a temperature parameter. Training a neural network to minimize this loss results in a feature space $z$ where transformed views of a particular image are localized under a particular similarity metric. As a result, similar images are located closer together within this feature space \citep{van2020scan}.

One of the key challenges with contrastive learning methods is choosing the right transformation for specific tasks. For example, using color distortion can remove information about the colors of specific flower species, which could result in poorer performance for these problems \citep{zhang2022rethinking}. Additionally, contrastive methods can perform poorly when the task of interest is not determined by the dominant features of an image \citep{chen2021intriguing}.

In practice, a neural network is trained with SimCLR using a projection head. The outputs of a backbone network $f$ (such as a ResNet) are passed through a multi-layer perceptron network $g$, so that $z_i = g(f(x'_i))$. When training the network for downstream tasks, this head can be discarded, or the first layer can be retained to improve performance in some cases \citep{chen2020big}.

\subsection{Semi-supervised learning}

Semi-supervised learning methods aim to improve model performance by using both labelled and unlabelled data during training. The key idea is that the underlying geometry of the dataset, which may be sparsely sampled by the labelled data, can be leveraged to push decision boundaries into low-density regions and improve performance beyond just using the labelled data \citep{van2020survey}. 

Pseudo-labelling \citep{lee2013pseudo} is an early semi-supervised technique where a model trained on a subset of the data is used to create pseudo-labels for the entire dataset. This technique has since been refined in various ways and can be highly effective \textemdash{} Fixmatch \citep{sohn2020fixmatch} is one such method that achieved state of the art on a number semi-supervised benchmarks when it was published, through clever application of image augmentations.


In this paper, we utilize PAWS (Predicting View-Assignments with Support Samples) \citep{assran2021semi}, a recent semi-supervised learning method that incorporates ideas from contrastive learning and avoids the explicit use of pseudo-labels. PAWS generates a similarity classifier from labelled support samples $\mathbf{z_s}$, using a similar form to Equation \ref{eq:NT-Xent}:

\begin{align}
    \phi_d(z_i,\mathbf{z_s})  = \sum_{(z_{s_j}, y_j) \in \mathbf{z_s}} \lb( \frac{e^{\delta(z_i, z_{s_j})/T}}{\sum_{z_{s_k} \in \mathbf{z_s}} e^{\delta(z_i, z_{s_k})/T}} \rb) y_j
    \label{eq:paws}
\end{align}

To train the neural network in a semi-supervised fashion, PAWS minimizes the cross-entropy $H(\cdot,\cdot)$ between different transformations of the same image, to encourage their predictions to be similar. Sharpening is also used to prevent the network collapsing to a trivial solution and make predictions more confident.

Many self-supervised methods are also strong semi-supervised approaches. In these methods, all of the data is used to build the representation, and the smaller labelled set is used to finetune the networks \citep{chen2020simple, caron2021emerging}. Additional training using distilation methods on the full dataset can further improve performance in some cases \citep{chen2020big}.

\subsection{Active Learning}

Active learning methods aim to identify the most optimal next set of data points to label based on an initial labelled set. There are two common strategies to approach this problem; either leveraging uncertainty information to select examples in close proximity to decision boundaries to improve them, or attempting to sample points within the dataset such that they are more representative \citep{budd2021survey}.

The core-set method \citep{sener2017active} is one such representational approach. This method is based on the premise that a sufficiently smooth loss function over a dataset can be bounded by using just a subset of the data points, and that convolutional neural networks can satisfy this property with certain loss functions. Given a set of spheres centered on the subset of selected points, the quality of this bound depends upon the radius required to cover all of the points within the dataset.

This naturally leads to an active learning scheme that selects points in order to minimize the $k$-centre optimization objective. The $k$-centre problem \citep{dyer1985simple} is a classic np-hard problem in computational geometry, where given a set of points $P$, the objective is to find a subset $X_n \subset P$ of $n$ points that minimises the objective function $\phi_{mM}(X_n)$

\begin{align}
    \phi_{mM}(X_n) = \max_{p \in P} \min_{x\in X_n} \delta(p, x)
    \label{eq:active_learning_coreset}
\end{align}

where $\delta$ is a particular metric \citep{dyer1985simple}. The core-set approach starts by training a neural network on a small subset of labelled points, then uses the last layer of the network before the classification head to project each image within the dataset onto a feature space $z$. This feature space is then used to determine the euclidean distance between images in the dataset, and new points are selected that minimise the objective function in Equation \ref{eq:active_learning_coreset}. This process can then iterate, with a new set of points being selected each time \citep{sener2017active}.

\subsection{Manifold learning}

Manifold learning, also referred to as non-linear dimensionality reduction, encompasses a range of techniques that project high-dimensional data onto lower-dimensional latent manifolds \citep{turaga2020manifold}. One objective of these techniques is to position items with similar features in close proximity within this latent space. They are frequently employed for the purpose of visualizing high-dimensional data; t-distributed stochastic neighbor embedding ($t$-SNE) is a popular approach for achieving this goal \citep{van2008visualizing}.

The key idea behind $t$-SNE is to create a probability distribution over pairs of points within a dataset that assigns a high probability to points $i,j$ that are close together, and a lower probability to points which are further away according to a particular metric $\delta$. A similar distribution can then be created for a simulated set of points within a lower dimensional space, and by minimising the Kullback-Leibler divergence between these the high and low dimensional distributions a lower dimensional embedding of the data can be created that corresponds to its higher dimensional counterpart \citep{van2008visualizing}. The probability distribution in the higher dimensional space is obtained from the conditional distribution

\begin{align}
    p_{j|i} = \frac{\exp(-\delta(x_i, x_j)/2\sigma_i^2)}{\sum_{k\not=i} \exp(-\delta(x_i, x_k)/2\sigma_i^2)}
\end{align}

where the bandwidth of the Gaussian kernels $\sigma_i$ is calculated for each datapoint so that the entropy of the conditional distribution for a particular point equals a predefined constant. As a result smaller values of $\sigma_i$ are used in denser regions of the dataset, and the effective number of nearest neighbours for each point is fixed.

Techniques such as SimCLR and other self-supervised learning methods can also be seen as manifold learning techniques, as they can be used to project images into a lower dimensional feature space that preserves relationships between images \citep{balestriero2022contrastive}.


\section{Methodology}
\subsection{Problem specification}

We consider a large dataset of images $D = \{x_i\}_{i\in[1,N]}$. We also consider a labelled test dataset that has been set aside, $D^* = \{x^*_i, y^*_i\}_{i\in[1,N^*]}$, which is used to test a model $f$, defined by parameters $\theta$. The aim is to identify the best set of training examples $X_n$ from $D$, given we may obtain at most $n$ labels to create a labelled subset $D_{X_n} = \{x'_{i}, y'_{i}\}_{x'_{i} \in X_n, i\in[1,n]}$. The best set of training examples are such that the model parameters $\theta(D, X_n)$, maximise

\begin{align}
    \text{Metric}(y^*, f(x^*|\theta(D, X_n)))
\end{align}

where $\text{Metric}$ defines the objective that measures performance on the test data $D^*$. Accuracy is a commonly used metric, but others, such as average precision or the F1 metric could also be used. 

Solving this equation for $X_n$ to find the optimal set $X^*_n$ is an np-hard problem. There are ${}_{N}C_n$ potential combinations of data instances that could be chosen, which increases rapidly for large $N$. This is a challenge also observed in active learning \citep{houlsby2011bayesian}. The function $\theta(D,X_n)$ could be defined by simply fine-tuning the model on the labelled subset $D_{X_n}$. However, better performance may be obtainable by using semi-supervised approaches, which will use all of the dataset $D$. In either case, not only does $X_n$ occupy a large space of potential solutions, but each evaluation of $\theta(D,X_n)$ will be computationally expensive. 

\subsection{A synergistic cold-start approach}

We propose to combine ideas from self-supervised and active learning approaches to address the cold-start learning problem, and test the performance of our approach in the context of semi-supervised learning. Our framework can be broken down into three steps

\begin{enumerate}
    \item Train a neural network using a self-supervised learning method.
    \item Use this neural network to encode the dataset into a feature space $z$, from which points can be selected for labelling using strategies similar to the core-set approach \citep{sener2017active}.
    \item Use semi-supervised learning, with the self-supervised weights as a starting point, to train a model using both the labelled and unlabelled data.
\end{enumerate}

The first and last step are well defined in the literature. There are many self-supervised and semi-supervised learning approaches that perform exceptionally well, and they are often combined to obtain state of the art performance \citep{cai2022semi}. However, much less is known about the second step and there are a number of unknowns. In particular, how do we define the space $z$ to get the best results? Or, how should we select points in an optimal manner?

While the original authors of the core-set approach suggest selecting points on the basis of solving the $k$-centre problem, this was in the context of approximating the loss over a dataset \textemdash{} when using self-supervised approaches to define a feature space, this may not necessarily be the best strategy.

\subsection{Choosing a feature space}

The core-set approach used the Euclidean distance between activations of the final fully connected layer of the neural network to determine the distance between two images \citep{sener2017active}. However, this choice may not be optimal. Specifically, SimCLR uses the cosine similarity between ouput vectors as the basis of measuring their comparability \citep{chen2020simple}. There are also different layers we can choose to define the feature space \textemdash{} we could choose the output of the backbone network, the projection head, or one of the intermediate layers. It may also be possible that these feature spaces need to be transformed to obtain the best performance, and manifold learning techniques such as $t$-SNE may improve results \citep{van2008visualizing}.

\subsection{A point-set selection strategy}

We initially looked to start with the approach of the core-set active learning paper \citep{sener2017active} for the cold learning case. The authors of this work used a mixed integer optimisation routine to find a solution to the greedy $k$-center problem, using the initially labelled points as a starting condition. Unfortunately, this approach did not converge in a reasonable amount of time when considering small numbers of labels without any initial selections, as detailed in the supporting information. 

Instead, we developed a fast and scalable approach based on a greedy starting point, in the spirit of this previous work. While we will leave consideration of these stategies in full to an ablation study in the supporting information (see section \ref{sec:ablation_space_filling_design}), we will refer to the greedy (mini-max) strategy in the rest of the paper. This most closely corresponded to the core-set approach as previously employed.

We compared the performance of this method to $k$-mediods, one of the simplest approaches to selecting points from the feature space $z$. This method is similar to $k$-means, but the centre of each cluster is given by a point within the dataset \citep{park2009simple}. The optimisation objective in this case is given by minimizing

\begin{align}
    \phi_{k\text{-mediods}}(X_n) = \sum_{p \in P} \min_{x_i\in X_n} \delta(x_i, p)
\end{align}

This was also considered previously, but it was found to not perform as well as the greedy $k$-center approach on the CIFAR-10 dataset \citep{sener2017active}.

\subsection{Evaluating performance}

If we focus solely on categorical classification problems, it becomes clear that in order for a labelled subset $X_n$ to perform well, it is necessary to include at least one image from each class. Without this, or any additional assumptions on the relationships between classes, \textit{empty} classes cannot be learned, leading to a decrease in the maximum achievable performance. For certain problems, identifying just one image from each class may suffice \citep{wang2022freematch}, while for more challenging problems, or those with a high degree of diversity within each class, it becomes crucial to sample the different features within a class appropriately.

We follow this logic to determine the most successful label-selection strategies from among those considered. First, we identify the best strategy for solving an unsupervised class discovery problem across a number of publicly available computer vision datasets (section \ref{sec:unsupervised_class_discovery_performance}). Then, we compare the performance of models fit using these \textit{smart} labelled subsets against those that use randomly selected samples (section \ref{sec:cold_start_learning_performance}). Across all our experiments, we maintain consistency in the hyperparameters employed across different datasets, with only a few exceptions. Additional details on the code and training approaches used can be found in the supporting information (section  \ref{sec:code_and_training}).

\subsection{Datasets}

We examined four image datasets; CIFAR-10 \citep{krizhevsky2009learning}, Imagenette \citep{imagenette}, EuroSAT \citep{helber2019eurosat}, and DeepWeeds \citep{olsen2019deepweeds}. We opted for Imagenette, a small subset of Imagenet \citep{deng2009imagenet} consisting of only 10 classes, as it allowed us to conduct experiments on Imagenet-like images in a manner feasible with the computational resources available. These datasets cover a range of image resolutions and classification problems, from identifying objects in scenes (CIFAR-10, Imagenette), classifying land use within a satellite image (EuroSAT), to identifying weeds within a natural environment (DeepWeeds). The latter two datasets also provided examples of cases where the classes within the data are not perfectly balanced, and in these cases the test sets were chosen randomly stratified by class. A summary of the classes in each dataset, and the number of testing and training images is provided in Table \ref{tbl:dataset_description} in the supporting information.

\section{Results}

\subsection{Unsupervised class discovery performance}
\label{sec:unsupervised_class_discovery_performance}

\begin{table*}[tp]
\centering
\caption{Ranking of each label selection strategy, scored using the proportion of runs where all classes are selected atleast once across the budgets considered. For each dataset we explored budgets with a multiple of the number of unique classes, up to 100 labels in total. Each case was run 20 times using a different starting seed for every dataset and budget combination.}
\label{tbl:unsupervised_class_detection_strategies}
\begin{tabular}{lllll}
  \hline
Method & \multicolumn{3}{l}{Components}  & Average score \\ 
\cmidrule{2-4} 
 & Metric & Transform & Layer  &  \\ 
  \hline
random (class balanced) &  &  &  & 1.00 \\ 
  kmediods & cosine distance & t-SNE & backbone &  $0.90^1$ \\ 
  kmediods & euclidean & t-SNE & backbone &  0.89 \\ 
  kmediods & euclidean & t-SNE & projection head &  0.87 \\ 
  kmediods & cosine distance &  & backbone &  0.86 \\ 
  kmediods & cosine distance & t-SNE & projection head &  0.84 \\ 
  kmediods & cosine distance &  & projection head &  0.81 \\ 
  kmediods & euclidean &  & backbone &  0.78 \\ 
  kmediods & euclidean &  & projection head & 0.75 \\ 
  greedy (mini-max) & euclidean &  & backbone & $0.68^2$ \\ 
  random &  &  &  & 0.67 \\
   \hline
\end{tabular}
\footnotesize

$\phantom{t}^1$ best $k$-mediods ($t$-SNE) $\phantom{t}^2$ best greedy (mini-max)

\end{table*}

\begin{figure*}[tp]
\centering
\includegraphics[width=1\textwidth]{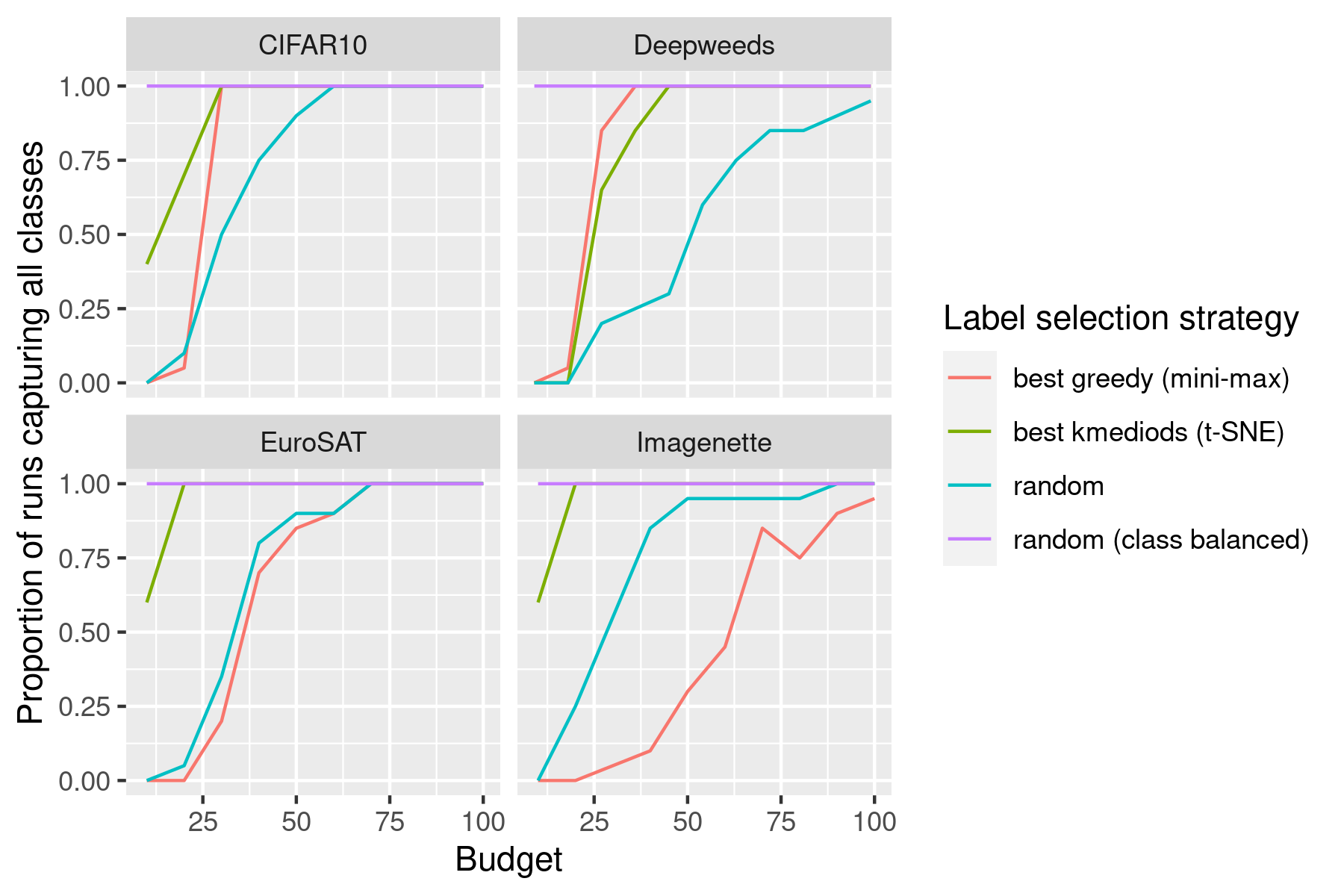}
\caption{Visualisation of unsupervised class detection performance versus labelling budget size, for a select number of strategies from Table \ref{tbl:unsupervised_class_detection_strategies}. The proportion for each budget-dataset combination is calculated from 20 runs. }
\label{fig:unsupervised_class_detection_strategies_figure}
\end{figure*}

In the first instance, we explored the performance of different cold-start strategies on an unsupervised class discovery problem. In this problem, we attempt to select an image of each class with the smallest possible dataset. We trained a network using SimCLR for each dataset, and selected images to be labelled using the resulting feature embeddings with a variety of different strategies. Further details on this process is provided in the supporting information (section \ref{sec:appendix_cold_start_strategies}).

We compared the performance of different strategies by looking at the proportion of runs for a particular budget that would sample at least one image from each class. We explored budgets for each dataset of multiples of the number of unique classes, up until 100 images. The average proportion of runs sampling all classes at least once was then averaged across different budgets and datasets to obtain a score for each strategy, which is shown in Table \ref{tbl:unsupervised_class_detection_strategies}.

It was found that the best label selection strategy used the $k$-mediod method and a $t$-SNE transform of the outputs of the SimCLR backbone. In the case that a cosine distance metric is used, for the particular budgets and datasets considered, 90\% of runs resulted in at least one image of each class being selected. This strategy will be referred to in the rest of this paper as the best $k$-mediods ($t$-SNE) strategy, and the results for each budget and dataset using this strategy are shown in Figure \ref{fig:unsupervised_class_detection_strategies_figure}. 

This $k$-mediods strategy performed much better than randomly selecting images, where only 67\% of runs sampled every class. It was also noted that this strategy performed better than the one that most closely corresponded to the approach taken in the core-set active learning approach \citep{sener2017active} \textemdash{} using the greedy (mini-max) method without the $t$-SNE transform, with a euclidean distance metric and on the backbone encodings (68\% of runs sampled every class). We will refer to this strategy as the best greedy (mini-max) strategy as it performed better than other combinations which used the greedy (mini-max) method, without the $t$-SNE transform. The results for each dataset for this method is also visualised in Figure \ref{fig:unsupervised_class_detection_strategies_figure}.

An ablation study of further label selection strategies, $t$-SNE clustering diagrams and visualisations of these image sets can be found in the supporting information.

\subsection{Cold-start learning performance}
\label{sec:cold_start_learning_performance}

\begin{figure*}[tp]
\centering
\includegraphics[width=1\textwidth]{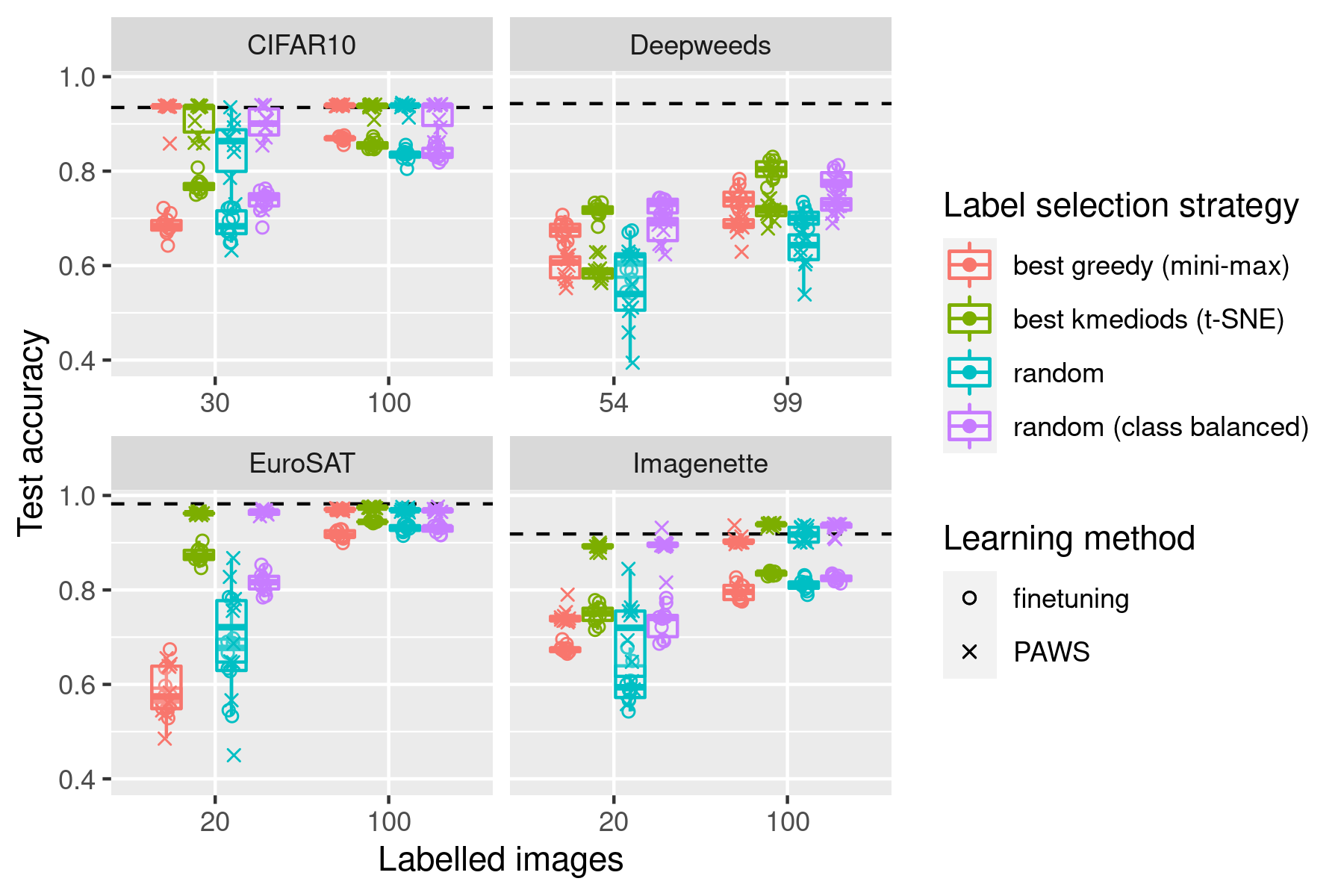}
\caption{Performance of finetuning and semi-supervised (PAWS) approaches using small numbers of labelled images, selected using different strategies. The dotted lines represent our fully supervised baselines for each dataset.}
\label{fig:results_all_finetuning_paws}
\end{figure*}

Although obtaining small subsets of images that effectively capture the different classes within a dataset is encouraging, it does not guarantee that these subsets will yield good results when used for training computer vision models. In this section, we investigate how the $k$-medoids and greedy (mini-max) strategies for selecting labelled subsets compare to randomly sampled sets of images in terms of their ability to train computer vision models.

We used two training approaches for downstream learning from the pretrained SimCLR weights \textemdash{} a supervised finetuning approach, inspired by semi-supervised SimCLR approaches \citep{chen2020big, chen2020simple}, and a PAWS semi-supervised approach \citep{assran2021semi}. We also considered using FixMatch, but found PAWS more consistent and much faster to fit. Further details on these methods are provided in the supporting information (section \ref{sec:appendix_classification_approaches}).

For illustrative purposes, we selected a budget for each dataset where all of the 20 subsets selected by the $k$-mediods strategy sampled atleast one image from each class present. We also selected a larger budget of 100 images, to explore how performance changed for each method in a higher label regime. As there were only nine classes present in the deepweeds dataset, we selected 99 images in this case. The results from training models based on these budget sizes are shown in Figure \ref{fig:results_all_finetuning_paws}. 

The performance of our $k$-mediods label selection strategy was promising, and could be considered superior to taking a random sample for all of the datasets and methods considered in the case of the small budget regime. In most cases, our $k$-mediods strategy performed similarly to the random (class balanced) case when training using PAWS, and in some cases did even better when finetuning.

The random sample selection strategy improves significantly in the higher label regime, and in most cases there is little difference in performance between the strategies when using PAWS. The primary exception is the DeepWeeds dataset \textemdash{} in this case, even with a larger sample, finetuning performance improves by around 10\% simply by using the $k$-mediods strategy over randomly sampling images. This likely reflects the imbalance within the dataset, which makes randomly sampling images less effective as rarer classes are less likely to be sampled.

In many cases the deficit in performance can be attributed to not all classes being sampled by the random strategy. For instance, in only six of ten cases did the random strategy sample all classes with 30 images for CIFAR10, and in no cases for Imagenette or EuroSAT for 20 images. Even in the case of 99 images on DeepWeeds, only nine of ten cases sampled all classes (see Table \ref{tbl:all_results_table} in the supporting information). However, even when all classes were sampled by the random strategy, there were still instances when performance was improved by using the $k$-mediods strategy, such as Imagenette with a budget of 100 images.

The greedy (mini-max) strategy had poorer performance in comparison to the $k$-mediods strategy, with the exception of on the CIFAR10 dataset. We obtained a median accuracy of 93.7\% (Table \ref{tbl:all_results_table}) in this case, which is competitive with the highest reported accuracy for this semi-supervised problem so far with 40 labels (95.1\%) \citep{wang2022freematch}. This is consistent with previous work \citep{sener2017active}, but seems anomalous when considering the performance of this method across the other datasets. This highlights the importance of using a diverse set of problems to benchmark new machine learning methods within this space.

\section{Discussion}

\begin{table}[t]
    \caption{A comparison to previous cold-start results in the literature on CIFAR-10.}
    \label{tbl:comparison_to_wang2022}
    \begin{tabular}{llllll}
      \hline
    Reference & \multicolumn{3}{l}{Components}   & Labels & Accuracy (\%) \\ 
    \cmidrule{2-4} 
     & Self-supervised & Selection strategy & Semi-supervised  & & \\ 
      \hline
    \citep{wang2022unsupervised} & SimCLR & USL & FixMatch & 40 & 90.4 \\ 
    \citep{wang2022unsupervised} & SimCLR & USL-T & FixMatch & 40 & 93.5 \\ 
      Ours & SimCLR & best $k$-mediods ($t$-SNE) & PAWS & 30 & \textbf{93.6}* \\
      \hline
    \citep{wang2022unsupervised} & SimCLR & USL & FixMatch & 100 & 93.2 \\ 
      Ours & SimCLR & best $k$-mediods ($t$-SNE) & PAWS & 100 & \textbf{93.9}* \\ 
       \hline
    \end{tabular}
    \footnotesize
    *We report our median accuracy here. We could not determine if \cite{wang2022unsupervised} selected a particular statistic to report performance.
\end{table}

While the cold-start problem has been previously considered in the computer vision field, as far as we are aware, \cite{wang2022unsupervised} provide the only other treatment of it that incorporates state-of-the-art self-supervised and semi-supervised approaches. A comparison of our methods with theirs is presented in Table \ref{tbl:comparison_to_wang2022}. Their best method, USL-T involves fine-tuning the self-supervised model which our methods do not require, making them simpler and more efficient while also obtaining slightly better results.

Similarly to improvements in semi-supervised learning, the outcomes of this research will make computer vision methods more accessible to researchers and others with large unlabelled datasets. It will also make computer vision work less labor dependent, and reduce reliance on crowd sourcing for annotations. By reducing the number of labels required for a high accuracy model, this work could also help make training sets more transparent, and reduce the quality control burden that goes with large annotated datasets.

In our experimentation, we found that the limitations of our approach were closely related to the current limitations of self-supervised learning. If the task of interest is not determined by the dominant features of an image, or is obscured by the transformations used to train the model, the self-supervised model may perform poorly at clustering the data into relevant groups \citep{chen2021intriguing}. In this case, the sampling strategies outlined here are unlikely to perform better than random sampling, and may even perform worse, as the clusters in the encoding space will not correspond to the classes of interest in the data.

When applying this method to new datasets, $t$-SNE plots of the self-supervised encodings can provide an indication as to whether our cold start learning approach will provide benefits. An example plot for each dataset considered here is shown in Figure \ref{fig:tsne_clusters} in the supporting information, for reference. If the $t$-SNE plot has well defined clusters that correspond to features of interest for classification, then it is more likely our approach will be successful.

\section{Conclusion}

In this paper, we demonstrate how well-established methods from manifold learning, clustering, and self-supervised learning can be combined to address the cold-start learning problem. We show that it is possible to select images from a large dataset \textit{a priori} that, when labeled, provide superior performance to a random sample across a range of different datasets using a consistent strategy in a simple yet effective manner.

Although we have used particular self-supervised and semi-supervised approaches in this paper to demonstrate the feasibility of this approach, it is possible that other methods could also be readily applied within the same framework. The ideas presented here to address the cold-start learning problem could also be adapted to other deep learning contexts, such as classifying texts or object detection. 

\pagebreak

\section{Acknowledgements}

We would like to thank Andrew Robinson for providing comment on the manuscript.
This research was undertaken using the LIEF HPC-GPGPU Facility hosted at the University of Melbourne. This Facility was established with the assistance of LIEF Grant LE170100200. 
This research/project was undertaken with the assistance of resources and services from the National Computational Infrastructure (NCI), which is supported by the Australian Government.


\bibliography{neurips_2023}
\vfill
\pagebreak
\appendix
\section{Ablation study: other point-selection strategies}
\label{sec:ablation_space_filling_design}

In this section, we describe alternative methods to the $k$-mediods approach employed in the paper, and examine their performance in greater detail. The $k$-center problem, as employed in the core-set approach \citep{sener2017active}, is an obvious choice, but we will also add an additional method inspired by similar work in the field of space-filling design \citep{pronzato2017minimax}. This field attempts to solve a similar problem \textemdash{} given a continuous domain of simulation inputs, how can we select points so that it is well-sampled? In this context, the $k$-centre problem is called the mini-max optimisation objective, where the maximum distance of any point to a selected label is minimised.

\begin{align}
    \phi_{mM}(X_n) = \max_{p \in P} \min_{x\in X_n} \delta(p, x)
\end{align}

We can also consider the maxi-min optimisation objective as an alternative choice, where the minimum distance between selected labels is maximised instead. This means maximising the value of $\phi_{Mm}(X_n)$;

\begin{align}
    \phi_{Mm}(X_n) &= \min_{x_i, x_j \in X_n, i \not= j} \delta(x_i, x_j)
\end{align}

Compared to the mini-max objective, this results in points being selected from the convex hull of the dataset, rather than from interior regions. In some cases where these points provide images that better capture the features present in the data, this might improve performance. Similarly, both of these methods look at optimising the maximum or minimum distances from a selected subset, while the $k$-mediods optimisation minimizes the sum of distances. This means they may be less sensitive to the density of the data compared to $k$-mediods, which could influence performance.

The maxi-min and mini-max optimisation objectives cannot be solved exactly in large systems, as they are within the class of np-hard problems \citep{cunningham1998combinatorial}. Instead, we start with the greedy algorithm \citep{dyer1985simple}, and we introduce novel approaches for modifying these greedy point sets in ways that further optimize them for either optimisation objective. This provides us with three methods to examine, the greedy approach (section \ref{sec:space_filling_design_greedy_k_centers}), the greedy (maxi-min) approach (section \ref{sec:space_filling_design_greedy_k_centers_maxi_min}), and the greedy (mini-max) approach (section \ref{sec:space_filling_design_greedy_k_centers_mini_max}). Benchmarking results for these methods are provided in section \ref{sec:space_filling_design_greedy_benchmarking}.

\subsection{Results}

\begin{table*}[tp]
\centering
\caption{Ranking of each label selection strategy, scored using the proportion of runs where all classes are selected atleast once across the budgets considered. For each dataset we explored budgets with value $xn < 100$, where $n$ is the number of unique classes in the dataset, and x is an integer. Each strategy was run 20 times using a different starting seed for every dataset and budget combination. The last column shows the average for each strategy over all of the datasets.}
\label{tbl:unsupervised_class_detection_strategies_all}
\tiny
\begin{tabular}{llll|rrrr|r}
  \hline
Method & \multicolumn{3}{l}{Components} & \multicolumn{4}{l}{Dataset score} & Average score \\ 
\cmidrule{2-4} \cmidrule(lr){5-8}
 & Metric & Transform & Layer & CIFAR10 & Deepweeds & EuroSAT & Imagenette &  \\ 
  \hline
random (class balanced) &  &  &  & 1.00 & 1.00 & 1.00 & 1.00 & 1.00 \\ 
  kmediods & cosine distance & t-SNE & backbone & 0.91 & 0.77 & 0.96 & 0.96 & 0.90 \\ 
  kmediods & euclidean & t-SNE & backbone & 0.92 & 0.79 & 0.92 & 0.94 & 0.89 \\ 
  kmediods & euclidean & t-SNE & projection head & 0.93 & 0.73 & 0.94 & 0.90 & 0.87 \\ 
  kmediods & cosine distance &  & backbone & 0.91 & 0.73 & 0.90 & 0.91 & 0.86 \\ 
  kmediods & cosine distance & t-SNE & projection head & 0.89 & 0.67 & 0.92 & 0.90 & 0.84 \\ 
  greedy (mini-max) & euclidean & t-SNE & backbone & 0.88 & 0.70 & 0.89 & 0.92 & 0.84 \\ 
  greedy (maxi-min) & euclidean & t-SNE & backbone & 0.84 & 0.74 & 0.89 & 0.90 & 0.84 \\ 
  greedy & euclidean & t-SNE & backbone & 0.86 & 0.77 & 0.87 & 0.86 & 0.84 \\ 
  greedy & cosine distance & t-SNE & backbone & 0.87 & 0.73 & 0.88 & 0.89 & 0.84 \\ 
  greedy (mini-max) & cosine distance & t-SNE & backbone & 0.88 & 0.71 & 0.88 & 0.88 & 0.84 \\ 
  greedy (maxi-min) & cosine distance &  & projection head & 0.88 & 0.70 & 0.82 & 0.89 & 0.82 \\ 
  greedy & cosine distance &  & projection head & 0.85 & 0.70 & 0.85 & 0.88 & 0.82 \\ 
  greedy (maxi-min) & cosine distance & t-SNE & backbone & 0.87 & 0.69 & 0.84 & 0.88 & 0.82 \\ 
  kmediods & cosine distance &  & projection head & 0.85 & 0.64 & 0.86 & 0.90 & 0.81 \\ 
  greedy (mini-max) & cosine distance & t-SNE & projection head & 0.88 & 0.62 & 0.87 & 0.88 & 0.81 \\ 
  greedy (mini-max) & euclidean & t-SNE & projection head & 0.86 & 0.61 & 0.86 & 0.90 & 0.81 \\ 
  greedy (maxi-min) & euclidean & t-SNE & projection head & 0.85 & 0.60 & 0.87 & 0.91 & 0.81 \\ 
  greedy & cosine distance & t-SNE & projection head & 0.87 & 0.62 & 0.87 & 0.85 & 0.80 \\ 
  greedy & cosine distance &  & backbone & 0.83 & 0.69 & 0.82 & 0.86 & 0.80 \\ 
  greedy (maxi-min) & cosine distance & t-SNE & projection head & 0.84 & 0.58 & 0.88 & 0.91 & 0.80 \\ 
  greedy & euclidean & t-SNE & projection head & 0.86 & 0.56 & 0.85 & 0.90 & 0.79 \\ 
  greedy (maxi-min) & cosine distance &  & backbone & 0.86 & 0.61 & 0.82 & 0.86 & 0.79 \\ 
  kmediods & euclidean &  & backbone & 0.79 & 0.69 & 0.83 & 0.81 & 0.78 \\ 
  kmediods & euclidean &  & projection head & 0.82 & 0.55 & 0.79 & 0.82 & 0.75 \\ 
  greedy (maxi-min) & euclidean &  & backbone & 0.84 & 0.76 & 0.79 & 0.58 & 0.74 \\ 
  greedy (maxi-min) & euclidean &  & projection head & 0.89 & 0.63 & 0.76 & 0.56 & 0.71 \\ 
  greedy (mini-max) & euclidean &  & backbone & 0.81 & 0.81 & 0.67 & 0.43 & 0.68 \\ 
  random &  &  &  & 0.72 & 0.51 & 0.70 & 0.74 & 0.67 \\ 
  greedy (mini-max) & euclidean &  & projection head & 0.78 & 0.59 & 0.77 & 0.47 & 0.65 \\ 
  greedy & euclidean &  & backbone & 0.80 & 0.69 & 0.58 & 0.40 & 0.62 \\ 
  greedy (mini-max) & cosine distance &  & backbone & 0.80 & 0.72 & 0.61 & 0.33 & 0.61 \\ 
  greedy (mini-max) & cosine distance &  & projection head & 0.74 & 0.58 & 0.76 & 0.34 & 0.61 \\ 
  greedy & euclidean &  & projection head & 0.80 & 0.54 & 0.69 & 0.37 & 0.60 \\ 
   \hline
\end{tabular}

\normalsize
\end{table*}

We followed the same procedure as in the results section of the paper to compare all of the different combinations of label selection methods discussed here and the other feature space components, such as choice of metric, transformation and neural network layer. The results are shown in Table \ref{tbl:unsupervised_class_detection_strategies_all}, including the results for the $k$-mediods method as well.

There are two things that appear to make a strategy more successful \textemdash{} using $k$-mediods or using a $t$-SNE transform. Using both together was found to be the most successful. For instance, the best greedy (mini-max) strategy acheived an average score of 0.84 with the $t$-SNE transform, and only 0.68 without. It is also noted that the cosine distance metric was much more successful overall in the absence of using a $t$-SNE transform on the encodings. This may reflect the fact that the SimCLR approach uses this metric as a basis for computing the loss during training. 

\begin{figure}[!b]
     \begin{subfigure}[b]{0.5\textwidth}
         \centering
         \includegraphics[width=\textwidth]{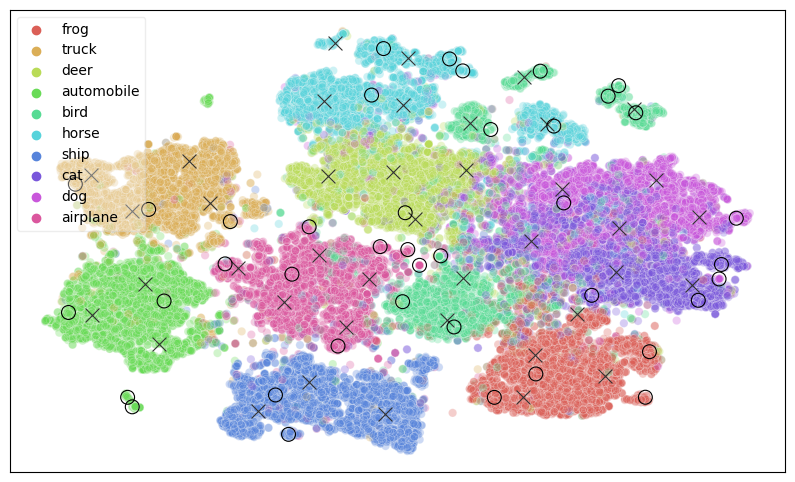}
         \caption{CIFAR-10 (40 labels)}
         \label{fig:tsne_clusters_cifar10}
     \end{subfigure}
     \hfill
     \begin{subfigure}[b]{0.5\textwidth}
         \centering
         \includegraphics[width=\textwidth]{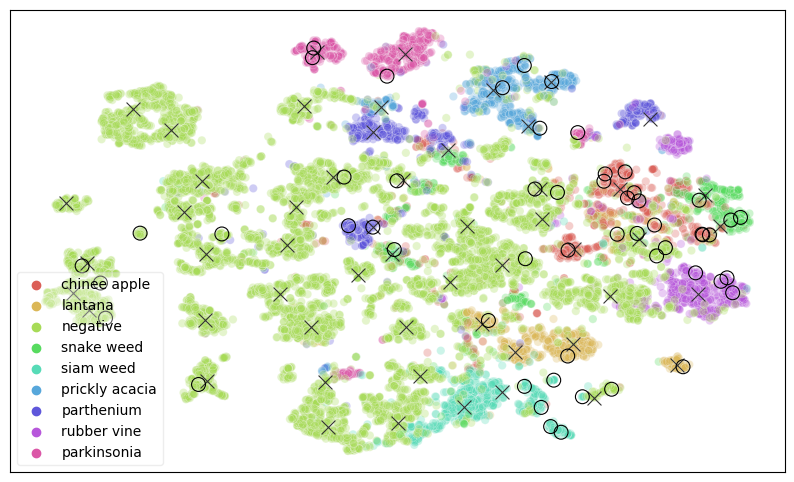}
         \caption{DeepWeeds (54 labels)}
         \label{fig:tsne_clusters_deepweeds}
     \end{subfigure}
     \begin{subfigure}[b]{0.5\textwidth}
         \centering
         \includegraphics[width=\textwidth]{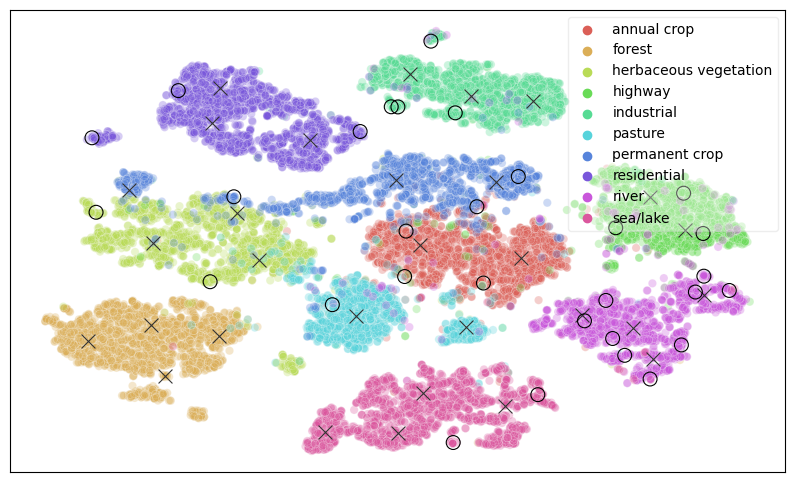}
         \caption{EuroSAT (30 labels)}
         \label{fig:tsne_clusters_eurosat}
     \end{subfigure}
     \hfill
     \begin{subfigure}[b]{0.5\textwidth}
         \centering
         \includegraphics[width=\textwidth]{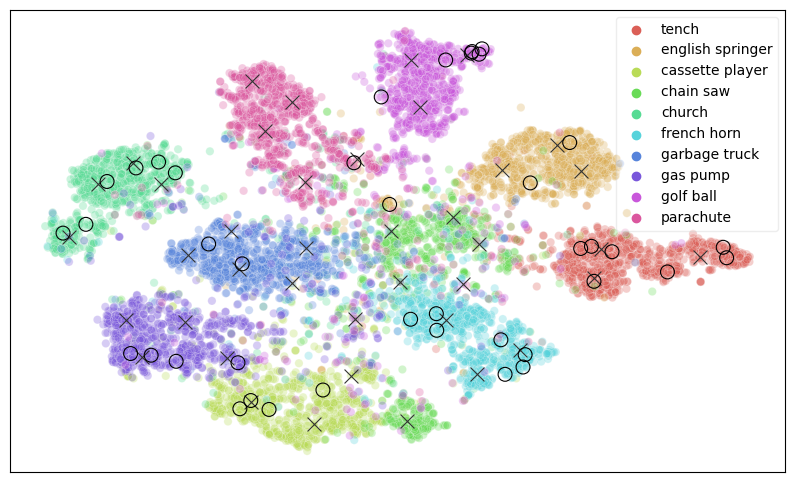}
         \caption{Imagenette (40 labels)}
         \label{fig:tsne_clusters_imagenette}
     \end{subfigure}
\caption{Selected images for labelling overlayed onto t-SNE clustering results for each dataset. The crosses show the points selected using a kmediods cosine distance t-SNE base encoder strategy, and the circles show the points selected using a greedy (maxi-min) euclidean backbone strategy.}
\label{fig:tsne_clusters}
\end{figure}

Under the euclidean metric, the volume occupied by different classes in the SimCLR feature space can be highly variable. It can be seen in Figure \ref{fig:tsne_clusters} that the non-transformed greedy (mini-max) strategy using the euclidean metric and backbone encodings avoids sampling particular classes in some datasets, such as chainsaws and parachutes for Imagenette, and forests in EuroSAT. This corresponds to worse performance for these datasets for this strategy in Table \ref{tbl:unsupervised_class_detection_strategies_all}.

The $t$-SNE transformation appears to improve the properties of the feature space for sampling, particularly when a euclidean metric is used. This is likely a result of how the embedding is created \textemdash{} by changing the bandwidth of the kernel so that each point has the same number of effective nearest neighbours. This would alleviate the imbalance in volume that the class clusters occupy in the transformed space.


\section{Space filling design: technical details}
\label{sec:space_filling_design}
\subsection{Algorithms for solving the greedy $k$-center problem}
\subsubsection{The greedy $k$-center algorithm}
\label{sec:space_filling_design_greedy_k_centers}

The greedy $k$-center algorithm \citep{dyer1985simple}, also known as the Gon algorithm \citep{garcia2019approximation}, was presented as a way to approximately solve the greedy $k$-center problem two-optimally, and in $O(Nn)$ time, where $N$ is the number of points in $P$ and $n$ is the desired subset size. Two-optimality requires that the approximation is within a factor of two of the optimal value.

The algorithm is based on the idea of always selecting the point furthest away from the currently chosen centers, and is shown in Algorithm \ref{alg:greedy_kcentre}. We note that this method is also two optimal for the maxi-min problem, in addition to the mini-max problem, and these proofs are provided in section \ref{sec:proofs_of_2_optimality}.

\begin{algorithm}
\caption{The greedy $k$-centre algorithm}\label{alg:greedy_kcentre}
\begin{algorithmic}
\Require $P = \{p_1,...,p_N\}$: set of points in $R^d$
\Require $n$: number of points to select from $P$ 
\Require $\delta$: a distance function defined in $R^d$
\State $c_1 = \text{randomly chosen index from 1 to the length of }P$
\State $C = \{ c_1 \}$
\For{$i=2$ to $n$}
    \State $c_i = \text{argmax}_k \min_{c_j \in C} \delta(p_k, p_{c_j})$ 
    \State $C = C \cup \{ c_i \}$
\EndFor
\State \textbf{return} $C$
\end{algorithmic}
\end{algorithm}

The solutions provided by the greedy algorithm can be made to be more optimal for either the maxi-min or mini-max optimisation problems with some simple steps. We chose to use these methods instead of others published in the literature, as they scaled well to high dimensions and large point sets.

\subsubsection{Mini-max optimisation}
\label{sec:space_filling_design_greedy_k_centers_mini_max}

For the mini-max case, we can start with the greedy solution and segment $P$ based on the closest center $x_i \in X_n$. This is equivalent to segmenting the space using a Voronoi partition. Then, we observe that a better solution to the global mini-max problem can always be found by solving a 1-centre mini-max problem locally in each of the resulting partitions. 

Done naively, this can be a slow and expensive process as it will scale as $O(nN^2)$, as the distance matrix inside each partition $K$ will need to be calculated. We can do much better than this by taking a subset of points $K^* \subset K$, and calculating the distance between the subset points $k_i^* \in K^*$ and all of the points $k_j \in K$ to choose a candidate center $c$. If we choose the point that minimises the distance to all points in $K^*$, we observe that this minimises the distance to all points in $K$ if the furthest point from $c$ in $K$ is also in $K^*$.

This can be seen by observing that 

\begin{align}
    \phi_{mM,1}^*(K) &= \min_{k_j \in K} \max_{k_i \in K} \delta(k_i, k_j)\\
    \phi_{mM,1}^*(K) &\ge  \min_{k_j \in K} \max_{k_i^* \in K^* \subset K} \delta(k_i^*, k_j)\\
    \phi_{mM,1}^*(K) &\le \max_{k_i \in K} \delta(k_i, c) \text{ for all } c \in K
\end{align}

This means that a subset of points $K^*$ can be used to find a lower bound to the optimal solution, which can then be cheaply verified by checking if the maximum distance from the candidate center $c$ to all the other points in $K$ is equal to this bound. 

We choose our subset $K^*$ by selecting points that had a maximal or minimal value in any dimension, and when a candidate center is found to be incorrect, the furthest away point from the candidate center is be added to $K^*$ and this process iterates until a solution is found.

We can repeat this for each partition until the quality $\phi_{mM}(X_n)$ of the solution no longer improves. This often resulted in a number of partitions $K$ of $P$ where all of the points within them were within $\phi_{mM}(X_n)$ of a different center. In these cases, these centers can be discard and re-assigned using the greedy approach with no loss to the quality of the point set. This algorithm is summarised in Algorithm \ref{alg:greedy_kcentre_minimax}.

\begin{algorithm}
\caption{The mini-max modified greedy $k$-centre algorithm \textemdash{} greedy (mini-max)}\label{alg:greedy_kcentre_minimax}
\begin{algorithmic}
\Require $P = \{p_1,...,p_m\}$: set of points in $R^d$
\Require $n$: number of points to select from $P$ 
\Require $\delta$: a distance function defined in $R^d$
\Require thresh: threshold at which center is redundant
\State Obtain $C$ by applying Algorithm \ref{alg:greedy_kcentre}.
\While{}
    \State $C^* = \{ \}$
    \For{$i=1$ to $n$}
        \State $K = \{ p_l \in P \text{ if } c_i = \text{argmin}_{c_j \in C} \delta(c_j, p_l) \}$
        \State $c^*_i = \text{argmin}_{k_j \in K} \max_{k_i \in K} \delta(k_i, k_j)$
        \State $C^* = C^* \cup \{ c^*_i \}$
    \EndFor
    \If{$\phi_{mM}(C) = \phi_{mM}(C^*)$}
        \State \textbf{break}
    \Else
        \State $C = C^*$
    \EndIf
\EndWhile
\For{$i=1$ to $n$}
    \State $K = \{ p_l \in P \text{ if } c_i = \text{argmin}_{c_j \in C} \delta(c_j, p_l) \}$
    \State $\tau = \text{mean}_{k_i \in K} \text{sum}_{c_j \in C, c_j \not= c_i} (\delta(k_i, c_j) \le  \phi_{mM}(C))$
    \State If $\tau > \text{thresh}$, delete $c_i$ from $C$
\EndFor
\State Use Algorithm \ref{alg:greedy_kcentre} to add further points to $C$ if required
\State Repeat previous while loop
\State \textbf{return} $C$
\end{algorithmic}
\end{algorithm}

\subsubsection{Maxi-min optimisation}
\label{sec:space_filling_design_greedy_k_centers_maxi_min}

A better solution for the maxi-min problem can also be obtained in a similar way. As the first point of a greedy solution is selected randomly, we can iterate through all of the selected points (starting with the first) and move them to a position that is farther away in each case, until we can no longer do so. This algorithm is summarised in Algorithm \ref{alg:greedy_kcentre_maximin}.

\begin{algorithm}
\caption{The maxi-min modified greedy $k$-centre algorithm \textemdash{} greedy (maxi-min)}\label{alg:greedy_kcentre_maximin}
\begin{algorithmic}
\Require $P = \{p_1,...,p_m\}$: set of points in $R^d$
\Require $n$: number of points to select from $P$ 
\Require $\delta$: a distance function defined in $R^d$
\State Obtain $C$ by applying Algorithm \ref{alg:greedy_kcentre}.
\While{}
    \State $C^* = \{ \}$
    \For{$i=1$ to $n$}
        \State $K = \{ p_l \in P \text{ if } c_i = \text{argmin}_{c_j \in C} \delta(c_j, p_l) \}$
        \State $c^*_i = \text{argmax}_{k_j \in K} \min_{c_l \in C, c_l \not= c_i} \delta(c_l, k_j)$
        \State $C^* = C^* \cup \{ c^*_i \}$
    \EndFor
    \If{$\phi_{Mm}(C) = \phi_{Mm}(C^*)$}
        \State \textbf{break}
    \Else
        \State $C = C^*$
    \EndIf
\EndWhile
\State \textbf{return} $C$
\end{algorithmic}
\end{algorithm}

\subsubsection{Benchmarking}
\label{sec:space_filling_design_greedy_benchmarking}

To demonstrate the effectiveness of these extensions to the greedy algorithm, we compare results with those published in the literature for the $k$-centre problem. These comparisons for a set of sample mini-max problems are shown in Table \ref{tbl:benchmarking_mini_max}. We were not able to find any comparisons for the maxi-min problem in this context, but we have compared the results from the greedy algorithm and our extension on these test instances in Table \ref{tbl:benchmarking_maxi_min}. These results show that significant improvements in the relevant metrics are obtained when using the greedy (maxi-min) or greedy (mini-max) algorithms, over just using the greedy algorithm alone, and that our greedy (maxi-min) algorithm is competitive with other polynomial time algorithms reported in the literature, particularly for large instances.

We observed that the approach used by \cite{sener2017active} can obtain the optimal answer for the smallest dataset, but was quite slow particularly for small values of $k$, and did not scale well to the larger datasets so results in these cases were not reported.

\begin{table*}[tp]
\centering
\caption{Value of $\phi_{mM}(X_n)$ obtained using various different strategies on TPSlib \citep{reinelt1991tsplib} benchmarking instances. The Gon, HS, Gon+, CDSh and CDSh+ strategies are polynomial-time results, BKS is a non-polynomial time algorithm and OPT are reported optimal values. These values have been reproduced here from the literature \citep{garcia2019approximation}. We report the method used in the coreset approach by \cite{sener2017active}, and then our results in the rightmost columns. The mean and standard deviation from ten runs of our greedy and greedy (mini-max) algorithms are shown. A threshold value of 0.9999 was used for the greedy (mini-max) algorithm. }
\label{tbl:benchmarking_mini_max}
\footnotesize
\begin{tabular}{lr|rr|rrrrr|r|rr}
  \hline
instance & k & OPT & BKS & Gon & HS & Gon+ & CDSh & CDSh+ & Coreset & greedy & greedy (mini-max) \\ 
  \hline
u1060 &  20 & 1581 &  & 2192 & 2140 & 1886 & 1780 & 1699 & * & 2109±87 & 1839±79 \\ 
  u1060 &  40 & 1021 &  & 1389 & 1360 & 1252 & 1218 & 1139 & 1020 & 1358±43 & 1268±41 \\ 
  u1060 &  60 & 781 &  & 1076 & 1152 & 985 & 944 & 906 & 786 & 1055±21 & 974±21 \\ 
  u1060 &  80 & 652 &  & 906 & 961 & 807 & 752 & 721 & 652 & 878±24 & 815±23 \\ 
  u1060 & 100 & 570 &  & 721 & 828 & 707 & 652 & 633 & 568 & 740±12 & 687±23 \\ 
  sw24978 &  25 &  & 1329 & 1618 & 1835 & 1550 & 1635 &&  & 1757±70 & 1446±96 \\ 
  sw24978 &  50 &  & 926 & 1240 & 1227 & 1088 & 1099 &&  & 1181±24 & 1015±51 \\ 
  sw24978 &  75 &  & 759 & 926 & 957 & 862 & 886 &&  & 925±17 & 818±23 \\ 
  sw24978 & 100 &  & 686 & 798 & 870 & 742 & 753 &&  & 784±15 & 713±16 \\ 
  bm33708 &  25 &  & 1184 & 1541 & 1688 & 1414 & 1491 &&  & 1598±62 & 1260±46 \\ 
  bm33708 &  50 &  & 823 & 1071 & 1191 & 951 & 981 &&  & 1054±28 & 841±18 \\ 
  bm33708 &  75 &  & 684 & 819 & 945 & 767 & 808 &&  & 825±10 & 694±19 \\ 
  bm33708 &  75 &  & 593 & 693 & 745 & 649 & 681 &&  & 825±10 & 694±19 \\ 
  ch71009 &  25 &  & 4429 & 6296 & 6519 & 5608 & 5806 &&  & 6187±139 & 4880±186 \\ 
  ch71009 &  50 &  & 3108 & 4040 & 4019 & 3642 & 3649 &&  & 3996±87 & 3400±68 \\ 
  ch71009 &  75 &  & 2554 & 3163 & 3584 & 2986 & 3015 &&  & 3223±64 & 2777±86 \\ 
  ch71009 & 100 &  & 2169 & 2777 & 2929 & 2525 & 2602 &&  & 2698±54 & 2356±54 \\ 
   \hline
\end{tabular}
\footnotesize
*failed to converge after 10 minutes.
\normalsize
\end{table*}

\begin{table*}[tp]
\centering
\caption{Value of $\phi_{Mm}(X_n)$ obtained using various different strategies on TPSlib \citep{reinelt1991tsplib} benchmarking instances. The mean and standard deviation from ten runs of our greedy and greedy (maxi-min) algorithms are reported.}
\label{tbl:benchmarking_maxi_min}
\footnotesize
\begin{tabular}{{lr|rr}}
  \hline
instance & k & greedy & greedy (maxi-min) \\ 
  \hline
u1060 &  20 & 2212±73 & 2627±122 \\ 
  u1060 &  40 & 1395±44 & 1551±62 \\ 
  u1060 &  60 & 1073±21 & 1134±29 \\ 
  u1060 &  80 & 886±28 & 935±43 \\ 
  u1060 & 100 & 752±11 & 775±21 \\ 
  sw24708 &  25 & 1822±73 & 2134±86 \\ 
  sw24708 &  50 & 1198±21 & 1331±54 \\ 
  sw24708 &  75 & 934±14 & 1026±57 \\ 
  sw24708 & 100 & 788±14 & 867±28 \\ 
  bm33708 &  25 & 1631±76 & 1870±148 \\ 
  bm33708 &  50 & 1064±31 & 1206±42 \\ 
  bm33708 &  75 & 833±14 & 928±37 \\ 
  bm33708 & 100 & 708±14 & 786±25 \\ 
  ch71009 &  25 & 6363±160 & 7310±288 \\ 
  ch71009 &  50 & 4042±96 & 4612±171 \\ 
  ch71009 &  75 & 3247±57 & 3633±159 \\ 
  ch71009 & 100 & 2712±60 & 3026±132 \\ 
   \hline
\end{tabular}

\normalsize
\end{table*}


\subsection{Proofs of 2-optimality}
\label{sec:proofs_of_2_optimality}
To prove our claim of 2-optimally for the greedy $k$-center algorithm, we need to introduce some more precise notation, as used in previous work \citep{pronzato2017minimax}. Given a point set $X_n = \{x_1, ... , x_n\} \subset P = \{p_1, .... , p_n\} \subset R^d$, and a metric $\delta$, we want to define an optimisation problem for selecting $X_n$ such that the space $P$ is sampled in a sensible fashion. 

The mini-max formulation of this problem aims to find the set $X_n \subset P$ such that the maximum distance of any point in $P$ to the closest point in $X_n$ is minimised. Specifically, we are attempting to minimise the objective 

\begin{align}
\phi_{mM}(X_n) &= \max_{p \in P} \min_{x \in X_n} \delta(p, x) \\
\phi_{mM}(X_{mM,n}^*)  &= \phi_{mM,n}^*(P) = \min_{X_n \subset P}  \phi_{mM}(X_n)
\label{eq:mini-max_objective}
\end{align}

We could also consider the maxi-min objective, that aims to find a point set $X_n \subset P$ such that the minimum distance between points in this set is maximised.

\begin{align}
\phi_{Mm}(X_n) &= \min_{x_i, x_j \in X_n, i \not= j} \delta(x_i, x_j) \\
\phi_{Mm}(X_{Mm,n}^*)  &= \phi_{Mm,n}^*(P) = \max_{X_n \subset P}  \phi_{Mm}(X_n)
\end{align}

Where the mini-max objective attempts to find representative points that are close to other points, the maxi-min objective will sample more points from the convex hull of the point set $P$.

\subsubsection{Mini-max optimisation}

We wish to prove that a point set constructed using the greedy algorithm $G_n$ is 2-optimal for the mini-max optimisation problem. This means that $\frac{1}{2} \phi_{mM}(G_n) \le \phi_{mM,n}^*(P) \le \phi_{mM}(G_n)$.

Let us consider the optimal solution $X_{mM,n}^*$ for a particular budget, and the next greedy solution in the sequence $G_{n+1}$. There are more points in $G_{n+1}$ than $X_{mM,n}^*$, so by the pigeonhole principle, atleast two points $g_i$ and $g_j$ within $G_{n+1}$ must be closer to one point $s_l$ in $X_{mM,n}^*$, compared to the other points. 

By the triangle inequality, this means
\begin{align}
\delta(g_i, g_j) \le \delta(g_i, s_l) + \delta(s_l, g_j) \le 2 \phi_{mM,n}^*(P)
\end{align}

We also have that $\delta(g_i, g_j) \ge \phi_{mM}(G_n)$, as the $n+1$ point selected by the greedy solution is always at distance $\phi_{mM}(G_n)$ from the other centres. This provides that $\phi_{mM}(G_n) \le 2 \phi_{mM,n}^*(P)$, from which the above expression can be obtained. Therefore, the greedy algorithm provides a 2-optimal solution to the mini-max optimisation problem. This proof is well known and follows similar proofs in the literature \citep{garcia2019approximation}.

\subsubsection{Maxi-min optimisation}

We wish to prove that a point set constructed using the greedy algorithm $G_n$ is 2-optimal for the maxi-min optimisation problem. This means that $\phi_{Mm}(G_n) \le \phi_{Mm,n}^*(P) \le 2\phi_{Mm}(G_n)$.

We will prove this by contradiction. Let us assume that $\phi_{Mm,n}^*(P) > 2\phi_{Mm}(G_n)$. 

We observe that $P$ is covered by balls of radius $\phi_{Mm}(G_{n})$ centered on points from $G_{n-1}$. This is because the point that determines the value of the mini-max objective at the $n-1$ index determines the value of the maxi-min objective at $n$, so $\phi_{mM}(G_{n-1}) = \phi_{Mm}(G_{n})$.

This covering means that for each point in our optimal solution $z_j \in X_{Mm,n}^*$, there is some point $g_i \in G_{n-1}$ such that $\delta(g_i, z_j) \le \phi_{Mm}(G_{n})$. By our assumption above, there only may be one such $z_j$, as $\delta(z_j, z_k) \ge \phi_{Mm,n}^*(P) > 2\phi_{Mm}(G_n)$. However, by the pigeonhole principle, this is a contradiction, as there are $n$ points in $X_{Mm,n}^*$ and $n-1$ points in $G_{n-1}$. 

Therefore $\phi_{Mm,n}^*(P) \le 2\phi_{Mm}(G_n)$, and the greedy solution is 2-optimal for the maxi-min optimisation problem.


\section{Code and training strategy}
\label{sec:code_and_training}

This section details how we implemented our space-filling design and the self-supervised and semi-supervised learning methods used in this paper. A copy of our code is available on github at \url{github.com/emannix/cold-paws-simclr-and-paws-semi-supervised-learning} and \url{github.com/emannix/cold-paws-labelling-selection-strategies}.

\subsection{SimCLR}

We started with a \texttt{pytorch} \citep{paszke2019pytorch} SimCLR implementation (\url{github.com/AndrewAtanov/simclr-pytorch}), which we modified to produce equivalent results to the official GitHub implementation (\url{github.com/google-research/simclr}) \citep{chen2020simple, chen2020big}. Some of the minor differences found included the application of momentum, initialisation of the networks and differences in image augmentation results between \texttt{tensorflow} and \texttt{torchvision}. 

The hyperparameters that were used to fit a ResNet18 neural network using SimCLR on CIFAR10 are provided in Table \ref{tbl:simclr_cifar10_hyperparameters}. We tried to use the same hyperparameters as used in the second SimCLR paper for this dataset, but we could only find them partly detailed \citep{chen2020big} and filled in the blanks with those reported on the GitHub repository. We used \texttt{pytorch} ResNet18 network that was equivalent to that published in the official GitHub implementation.

\begin{table}[H]
\centering
\caption{Hyperparameters used for fitting SimCLR models to the CIFAR10 dataset.}
\label{tbl:simclr_cifar10_hyperparameters}
 \begin{tabular}{m{6cm} | m{6cm}} 
 \hline
 Hyperparameter & Value \\
 \hline
 Batch size & 1024 \\
 Learning rate & 3.2 \\ 
 Weight decay & 1e-4 \\
 Learning rate schedule &  Cosine decay with linear warmup \\
 Optimiser & LARS SGD \\
 Epochs & 800 \\
 Scheduler warm-up proportion & 0.05 \\
 Projection head layers & 2  \\
 Projection head output dimension & 128  \\
 Temperature & 0.2 \\
 \hline
 \end{tabular}
\end{table}

We also used the same image augmentation approach, but with \texttt{torchvision} instead of \texttt{tensorflow} augmentations as they were much faster within a \texttt{pytorch} training loop, although they did produce slightly different results. The only other modification we have purposely made to the original implementation is to normalise the images after the transformations, as is commonly done in most semi-supervised and supervised learning approaches. In our experiments this resulted in similar performance when the models were fine-tuned afterwards for the CIFAR10 dataset, compared to not normalising the images. The image augmentations used are summarised in Table \ref{tbl:simclr_cifar10_augmentations}.

\begin{table}[H]
\centering
\caption{Image augmentations used for fitting SimCLR models to the CIFAR10 dataset. The names of these augmentations relate to the relevant function in the \texttt{torchvision} package.}
\label{tbl:simclr_cifar10_augmentations}
 \begin{tabular}{m{6cm} m{6cm}} 
 \hline
 Augmentation & Parameters \\
 \hline
 \texttt{RandomResizedCrop} & size=32, scale=(0.08, 1.0), interpolation=bicubic \\
 \hline
 \texttt{RandomHorizontalFlip} & p=0.5 \\ 
 \hline
 \texttt{RandomApply}(\texttt{ColorJitter}) & brightness=0.4, contrast=0.4, saturation=0.4, hue=0.1, p=0.8 \\
 \hline
 \texttt{RandomGrayscale} & p=0.2 \\
 \hline
 \texttt{Normalize} & mean =(0.4914, 0.4822, 0.4465), sd=(0.2023, 0.1994, 0.2010) \\
 \hline
 \end{tabular}
\end{table}

The same parameters were also used for training ResNet18 models using SimCLR on the Imagenette, DeepWeeds and EuroSAT datasets. The only exceptions in these cases were that an image size of $224$ was used in the \texttt{RandomResizedCrop} function, and mean=(0.485, 0.456, 0.406) and sd=(0.229, 0.224, 0.225) was used for the \texttt{Normalize} parameters. We also only trained the Imagenette SimCLR model for 200 epochs, as we found this resulted in models that generalised slightly better.

\subsection{Cold-start strategies}
\label{sec:appendix_cold_start_strategies}

To obtain the feature space for our datasets, we took the outputs of our trained models from the base ResNet18 encoder, as well as the projection head that is used to evalutate the SimCLR loss. For CIFAR10, we simply normalised the validation images using the same normalisation function as in the previous section. For the images from Imagenette, DeepWeeds and EuroSAT, we resized the smaller edge of the image to 256 pixels, and cropped the centre 224x224 pixel square, as is standard practice. We then normalised these images using the same mean and standard deviations as done previously.

We used these feature spaces to select candidate sets of points to labels using 32 different strategies. These were obtained by using one of four different methods, a euclidean or cosine distance metric, the feature space defined by the projection head or base encoder, and applying or not applying $t$-SNE to reduce the dimensionality of the feature space to two dimensions. We supplemented these with two additional comparative strategies \textemdash{} a random strategy, where images are selected for labelling randomly, and a random (class balanced) strategy, where images are selected randomly stratified by class. 

The four methods used included the greedy strategy (Algorithm \ref{alg:greedy_kcentre}), the greedy (mini-max) strategy (Algorithm \ref{alg:greedy_kcentre_minimax}) and the greedy (maxi-min) strategy (Algorithm \ref{alg:greedy_kcentre_maximin}). These did not require any additional hyperparameters, except for the greedy (mini-max) strategy where we used a threshold value of 0.9999. The final method we tested was a $k$-mediods approach. We used the \texttt{sci-kit learn} implementation, with init=k-mediods++, method=alternate and max\_iter=1000 \citep{scikit-learn}.

We also used the \texttt{sci-kit learn} implementation of the $t$-SNE approach \citep{scikit-learn}. We used a perplexity of 40 and 1000 iterations, which was found to produce stable results for all of the datasets. As this process is stochastic, a different $t$-SNE encoding for the dataset was calculated for each iteration. When the cosine distance metric was used with $t$-SNE, this metric was used to fit the lower dimensional $t$-SNE space, and euclidean distance was used when applying the different label selection methods.

\subsection{Classification approaches}
\label{sec:appendix_classification_approaches}
\subsubsection{Finetuning}

We attempted to use the same approach as previously done in the SimCLR studies \citep{chen2020simple, chen2020big} for CIFAR10 to finetune our models on these labelled subsets. The hyper-parameters used are summarised in Table \ref{tbl:finetuning_cifar10_hyperparameters}. In this case, we did not have a fixed number of epochs. Rather, we trained our models until the test accuracy did not improve further, as there were large differences in the optimal number of training epochs depending on the size of the labelled subset and the particular dataset.

\begin{table}[H]
\centering
\caption{Hyperparameters used for fine-tuning models pre-trained using SimCLR.}
\label{tbl:finetuning_cifar10_hyperparameters}
 \begin{tabular}{m{6cm}  m{6cm}} 
 \hline
 Hyperparameter & Value \\
 \hline
 Batch size & 1024 \\
 Learning rate & 0.16 \\ 
 Weight decay & 0 \\
 Learning rate schedule &  Constant \\
 Optimiser & LARS SGD \\
 Epochs & X \\
 \hline
 \end{tabular}
\end{table}

We also used the same augmentation approach as the previous SimCLR papers \citep{chen2020simple, chen2020big}. This involved using the same augmentations in Table \ref{tbl:simclr_cifar10_augmentations}, but removing the color jitter and random greyscale functions. We also used these hyperparameters and augmentations for training our fully supervised baselines. These are not the strongest supervised baselines possible for the datasets we considered, but provided a useful comparison for our semi-supervised approaches.

In many cases we were fitting models using imbalanced training data. To correct for this, we used class balanced loss with a beta value of 0.9999 \citep{cui2019class}.

\subsubsection{PAWS}

We were able to incorporate the official PAWS \citep{assran2021semi} implementation provided on GitHub (\url{github.com/facebookresearch/suncet}) into our code-base. We used the same parameters as they did for CIFAR-10, with a few exceptions that are summarised in Table \ref{tbl:paws_cifar10_hyperparameters}. We used the exact same image augmentations as implemented in the PAWS repository.

\begin{table}[H]
\centering
\caption{Hyperparameters used for using PAWS to train neural networks.}
\label{tbl:paws_cifar10_hyperparameters}
 \begin{tabular}{m{6cm}  m{6cm}} 
 \hline
 Hyperparameter & Value \\
 \hline
 Weight decay & 1e-4 \\
 Supervised batch size & 160 \\ 
 Epochs & 200 \\
 \hline
 \end{tabular}
\end{table}

In adapting the hyperparameters and augmentations for CIFAR10 to the other datasets we considered, we used the same strategy as for SimCLR. The image size inputs to the random resized crops were changed, the normalisation constants were changed and the validation transformations were modified as previously described. However, as a number of different sized crops are used, we also used the Imagenet values for the scales of these.

The performance that we report for the PAWS method uses the nearest neighbour approach described within the paper \citep{assran2021semi} to classify images and report accuracy. We did not fine-tune these models further.

\section{Computation time}

The experiments outlined in this paper required approximately a thousand hours to complete with an NVIDIA Tesla Volta V100 GPU and twelve CPU cores. The majority of these hours were dedicated to fitting models using PAWS due to the required number of replicates. These runs took about three hours each, for most of the data sets.

\vfill
\FloatBarrier
\pagebreak
\section{Further figures and tables}

\begin{table}[H]
\centering
\caption{Summary of the datasets used in this paper, their classes, and the representation of these classes across the the training and test data.}
\label{tbl:dataset_description}
\tiny
\begin{tabular}{lrr|lrr|lrr|lrr}
  \hline
\multicolumn{3}{c|}{CIFAR10} & \multicolumn{3}{c|}{Imagenette} & \multicolumn{3}{c|}{EuroSAT}  & \multicolumn{3}{c}{DeepWeeds} \\
class & train & test & class & train & test & class & train & test & class & train & test \\ 
  \hline
airplane & 5000 & 1000 & tench & 963 & 387 & annual crop & 2500 & 500 & chinee apple & 825 & 300 \\ 
  automobile & 5000 & 1000 & english springer & 955 & 395 & forest & 2500 & 500 & lantana & 764 & 300 \\ 
  bird & 5000 & 1000 & cassette player & 993 & 357 & herbaceous vegetation & 2500 & 500 & negative & 8806 & 300 \\ 
  cat & 5000 & 1000 & chain saw & 858 & 386 & highway & 2000 & 500 & parkinsonia & 731 & 300 \\ 
  deer & 5000 & 1000 & church & 941 & 409 & industrial & 2000 & 500 & parthenium & 722 & 300 \\ 
  dog & 5000 & 1000 & french horn & 956 & 394 & pasture & 1500 & 500 & prickly acacia & 762 & 300 \\ 
  frog & 5000 & 1000 & garbage truck & 961 & 389 & permanent crop & 2000 & 500 & rubber vine & 709 & 300 \\ 
  horse & 5000 & 1000 & gas pump & 931 & 419 & residential & 2500 & 500 & siam weed & 774 & 300 \\ 
  ship & 5000 & 1000 & golf ball & 951 & 399 & river & 2000 & 500 & snake weed & 716 & 300 \\ 
  truck & 5000 & 1000 & parachute & 960 & 390 & sea/lake & 2500 & 500 &  &  &  \\ 
   \hline
\end{tabular}

\normalsize
\end{table}

\begin{figure}[p]
     \begin{subfigure}[b]{0.5\textwidth}
         \centering
         \includegraphics[width=\textwidth]{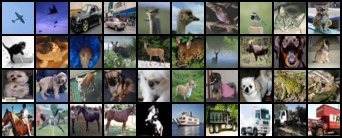}
         \caption{CIFAR-10 (40 labels) random}
         \label{fig:example_images_cifar10_random}
     \end{subfigure}
     \hfill
     \begin{subfigure}[b]{0.5\textwidth}
         \centering
         \includegraphics[width=\textwidth]{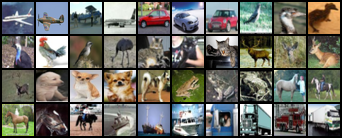}
         \caption{CIFAR-10 (40 labels) best $k$-mediods ($t$-SNE)}
         \label{fig:example_images_cifar10_kmediods}
     \end{subfigure}
     \begin{subfigure}[b]{0.5\textwidth}
         \centering
         \includegraphics[width=\textwidth]{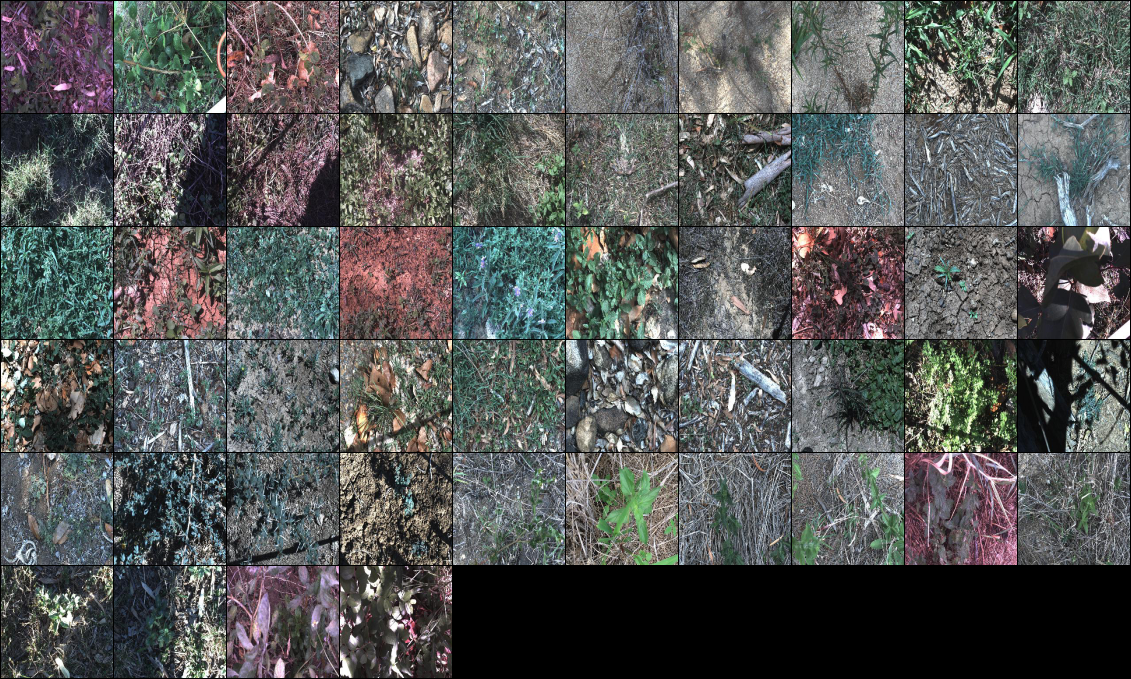}
         \caption{DeepWeeds (54 labels) random}
         \label{fig:example_images_deepweeds_random}
     \end{subfigure}
     \hfill
     \begin{subfigure}[b]{0.5\textwidth}
         \centering
         \includegraphics[width=\textwidth]{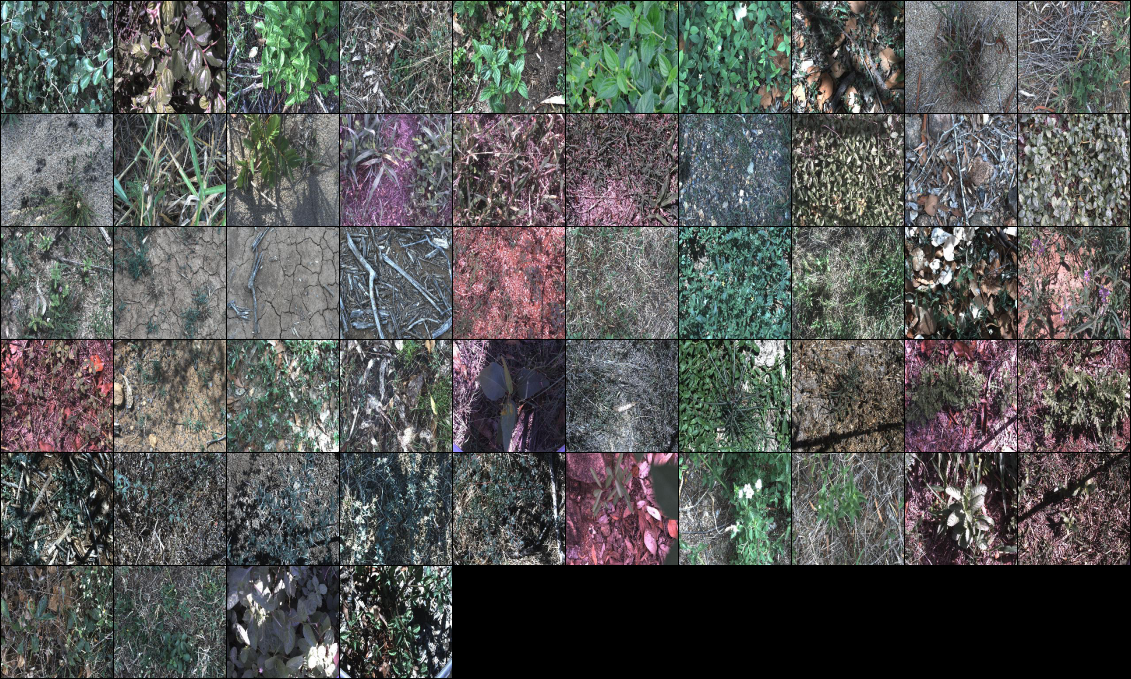}
         \caption{DeepWeeds (54 labels) best $k$-mediods ($t$-SNE)}
         \label{fig:example_images_deepweeds_kmediods}
     \end{subfigure}
     \begin{subfigure}[b]{0.5\textwidth}
         \centering
         \includegraphics[width=\textwidth]{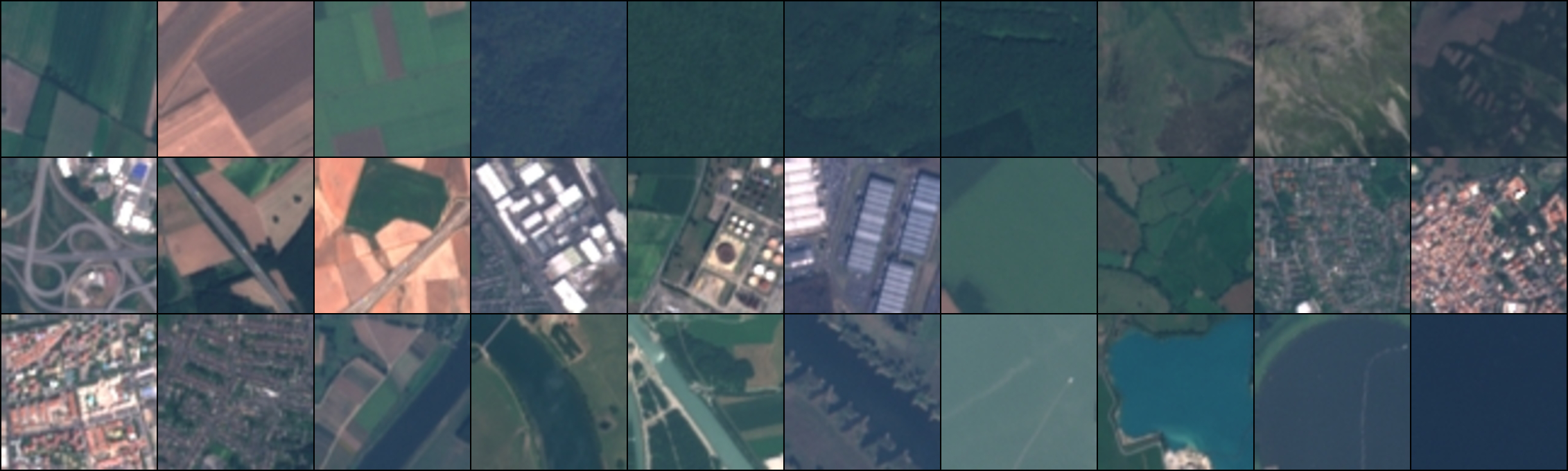}
         \caption{EuroSAT (30 labels) random}
         \label{fig:example_images_eurosat_random}
     \end{subfigure}
     \hfill
     \begin{subfigure}[b]{0.5\textwidth}
         \centering
         \includegraphics[width=\textwidth]{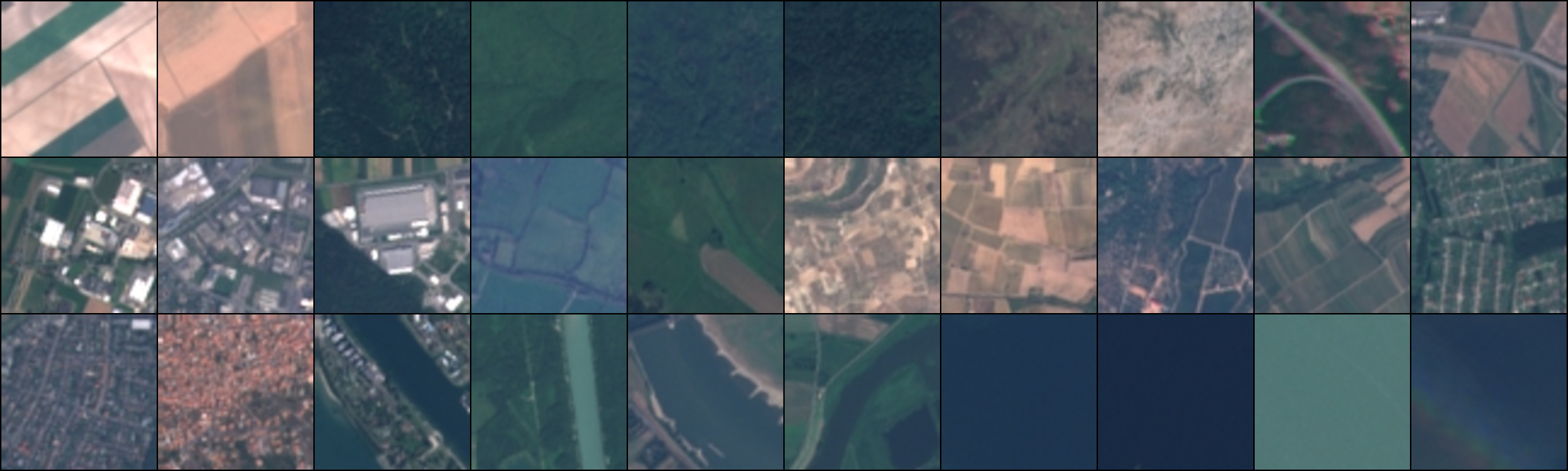}
         \caption{EuroSAT (30 labels) best $k$-mediods ($t$-SNE)}
         \label{fig:example_images_eurosat_kmediods}
     \end{subfigure}
     \begin{subfigure}[b]{0.5\textwidth}
         \centering
         \includegraphics[width=\textwidth]{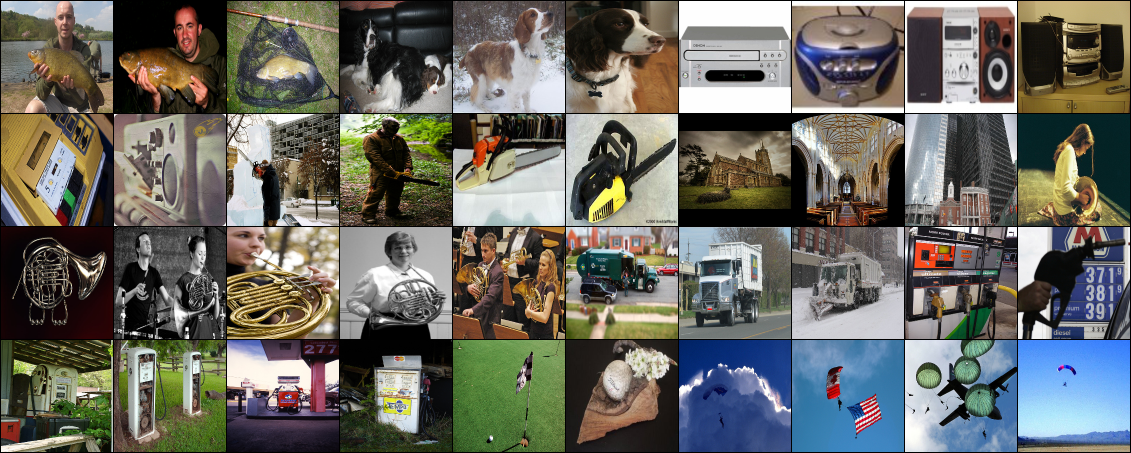}
         \caption{Imagenette (40 labels) random}
         \label{fig:example_images_imagenette_random}
     \end{subfigure}
     \hfill
     \begin{subfigure}[b]{0.5\textwidth}
         \centering
         \includegraphics[width=\textwidth]{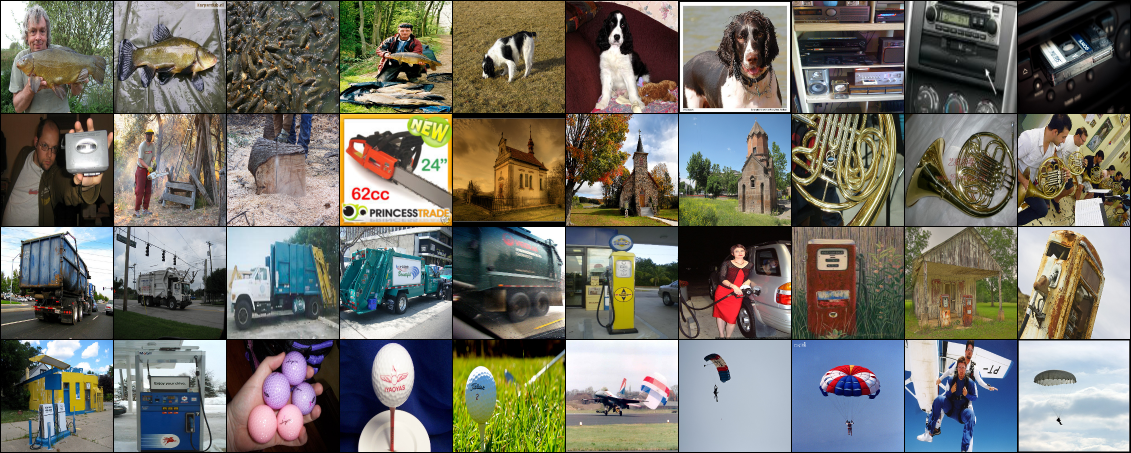}
         \caption{Imagenette (40 labels) best $k$-mediods ($t$-SNE)}
         \label{fig:example_images_imagenette_kmediods}
     \end{subfigure}
\caption{Example subsets of images selected for labelling from each dataset.}
\label{fig:example_images}
\end{figure}

\FloatBarrier

\begin{table}[p]
\centering
\caption{Table of results for the finetuning and PAWS fitting experiments. The last column shows then number of runs where atleast one image was present for each class.}
\label{tbl:all_results_table}
\tiny
\begin{tabular}{lllllrr}
  \hline
Dataset & Learning method & Number of labelled images & Label selection strategy & Test accuracy & Runs & (complete classes) \\ 
  \hline
CIFAR10 & finetuning & 30 & best greedy (mini-max) & 0.684 (0.642-0.722) &  10 &  10 \\ 
  CIFAR10 & finetuning & 30 & best kmediods (t-SNE) & 0.769 (0.750-0.808) &  10 &  10 \\ 
  CIFAR10 & finetuning & 30 & random & 0.683 (0.649-0.724) &  10 &   6 \\ 
  CIFAR10 & finetuning & 30 & random (class balanced) & 0.743 (0.681-0.763) &  10 &  10 \\ 
  CIFAR10 & finetuning & 100 & best greedy (mini-max) & 0.870 (0.855-0.876) &  10 &  10 \\ 
  CIFAR10 & finetuning & 100 & best kmediods (t-SNE) & 0.855 (0.846-0.874) &  10 &  10 \\ 
  CIFAR10 & finetuning & 100 & random & 0.836 (0.805-0.855) &  10 &  10 \\ 
  CIFAR10 & finetuning & 100 & random (class balanced) & 0.833 (0.818-0.860) &  10 &  10 \\ 
  CIFAR10 & PAWS & 30 & best greedy (mini-max) & 0.937 (0.858-0.939) &  10 &  10 \\ 
  CIFAR10 & PAWS & 30 & best kmediods (t-SNE) & 0.936 (0.858-0.939) &  10 &  10 \\ 
  CIFAR10 & PAWS & 30 & random & 0.864 (0.632-0.935) &  10 &   6 \\ 
  CIFAR10 & PAWS & 30 & random (class balanced) & 0.901 (0.718-0.940) &  10 &  10 \\ 
  CIFAR10 & PAWS & 100 & best greedy (mini-max) & 0.939 (0.938-0.941) &  10 &  10 \\ 
  CIFAR10 & PAWS & 100 & best kmediods (t-SNE) & 0.939 (0.909-0.942) &  10 &  10 \\ 
  CIFAR10 & PAWS & 100 & random & 0.939 (0.914-0.945) &  10 &  10 \\ 
  CIFAR10 & PAWS & 100 & random (class balanced) & 0.939 (0.861-0.942) &  10 &  10 \\ 
  CIFAR10 & supervised &  &  & 0.935 &   1 &  \\ 
  Deepweeds & finetuning & 54 & best greedy (mini-max) & 0.676 (0.642-0.707) &  10 &  10 \\ 
  Deepweeds & finetuning & 54 & best kmediods (t-SNE) & 0.719 (0.682-0.734) &  10 &  10 \\ 
  Deepweeds & finetuning & 54 & random & 0.604 (0.541-0.674) &  10 &   5 \\ 
  Deepweeds & finetuning & 54 & random (class balanced) & 0.731 (0.705-0.743) &  10 &  10 \\ 
  Deepweeds & finetuning & 99 & best greedy (mini-max) & 0.740 (0.684-0.783) &  10 &  10 \\ 
  Deepweeds & finetuning & 99 & best kmediods (t-SNE) & 0.804 (0.765-0.830) &  10 &  10 \\ 
  Deepweeds & finetuning & 99 & random & 0.701 (0.657-0.735) &  10 &   9 \\ 
  Deepweeds & finetuning & 99 & random (class balanced) & 0.777 (0.761-0.813) &  10 &  10 \\ 
  Deepweeds & PAWS & 54 & best greedy (mini-max) & 0.606 (0.552-0.660) &  10 &  10 \\ 
  Deepweeds & PAWS & 54 & best kmediods (t-SNE) & 0.586 (0.563-0.629) &  10 &  10 \\ 
  Deepweeds & PAWS & 54 & random & 0.540 (0.394-0.630) &  10 &   5 \\ 
  Deepweeds & PAWS & 54 & random (class balanced) & 0.690 (0.624-0.712) &  10 &  10 \\ 
  Deepweeds & PAWS & 99 & best greedy (mini-max) & 0.685 (0.629-0.740) &  10 &  10 \\ 
  Deepweeds & PAWS & 99 & best kmediods (t-SNE) & 0.717 (0.679-0.743) &  10 &  10 \\ 
  Deepweeds & PAWS & 99 & random & 0.643 (0.539-0.691) &  10 &   9 \\ 
  Deepweeds & PAWS & 99 & random (class balanced) & 0.731 (0.690-0.781) &  10 &  10 \\ 
  Deepweeds & supervised &  &  & 0.943 &   1 &  \\ 
  EuroSAT & finetuning & 20 & best greedy (mini-max) & 0.566 (0.528-0.674) &  10 &   0 \\ 
  EuroSAT & finetuning & 20 & best kmediods (t-SNE) & 0.872 (0.846-0.904) &  10 &  10 \\ 
  EuroSAT & finetuning & 20 & random & 0.677 (0.533-0.785) &  10 &   0 \\ 
  EuroSAT & finetuning & 20 & random (class balanced) & 0.816 (0.784-0.853) &  10 &  10 \\ 
  EuroSAT & finetuning & 100 & best greedy (mini-max) & 0.914 (0.899-0.928) &  10 &  10 \\ 
  EuroSAT & finetuning & 100 & best kmediods (t-SNE) & 0.944 (0.941-0.946) &  10 &  10 \\ 
  EuroSAT & finetuning & 100 & random & 0.931 (0.915-0.941) &  10 &  10 \\ 
  EuroSAT & finetuning & 100 & random (class balanced) & 0.933 (0.916-0.938) &  10 &  10 \\ 
  EuroSAT & PAWS & 20 & best greedy (mini-max) & 0.574 (0.485-0.655) &  10 &   0 \\ 
  EuroSAT & PAWS & 20 & best kmediods (t-SNE) & 0.962 (0.958-0.966) &  10 &  10 \\ 
  EuroSAT & PAWS & 20 & random & 0.721 (0.450-0.867) &  10 &   0 \\ 
  EuroSAT & PAWS & 20 & random (class balanced) & 0.966 (0.799-0.971) &  10 &  10 \\ 
  EuroSAT & PAWS & 100 & best greedy (mini-max) & 0.970 (0.966-0.973) &  10 &  10 \\ 
  EuroSAT & PAWS & 100 & best kmediods (t-SNE) & 0.974 (0.967-0.977) &  10 &  10 \\ 
  EuroSAT & PAWS & 100 & random & 0.969 (0.964-0.976) &  10 &  10 \\ 
  EuroSAT & PAWS & 100 & random (class balanced) & 0.969 (0.965-0.976) &  10 &  10 \\ 
  EuroSAT & supervised &  &  & 0.982 &   1 &  \\ 
  Imagenette & finetuning & 20 & best greedy (mini-max) & 0.674 (0.665-0.696) &  10 &   0 \\ 
  Imagenette & finetuning & 20 & best kmediods (t-SNE) & 0.750 (0.715-0.778) &  10 &  10 \\ 
  Imagenette & finetuning & 20 & random & 0.594 (0.543-0.678) &  10 &   0 \\ 
  Imagenette & finetuning & 20 & random (class balanced) & 0.741 (0.686-0.783) &  10 &  10 \\ 
  Imagenette & finetuning & 100 & best greedy (mini-max) & 0.796 (0.776-0.826) &  10 &  10 \\ 
  Imagenette & finetuning & 100 & best kmediods (t-SNE) & 0.837 (0.829-0.841) &  10 &  10 \\ 
  Imagenette & finetuning & 100 & random & 0.812 (0.790-0.831) &  10 &  10 \\ 
  Imagenette & finetuning & 100 & random (class balanced) & 0.825 (0.814-0.835) &  10 &  10 \\ 
  Imagenette & PAWS & 20 & best greedy (mini-max) & 0.738 (0.731-0.790) &  10 &   0 \\ 
  Imagenette & PAWS & 20 & best kmediods (t-SNE) & 0.892 (0.878-0.900) &  10 &  10 \\ 
  Imagenette & PAWS & 20 & random & 0.720 (0.558-0.845) &  10 &   0 \\ 
  Imagenette & PAWS & 20 & random (class balanced) & 0.895 (0.816-0.932) &  10 &  10 \\ 
  Imagenette & PAWS & 100 & best greedy (mini-max) & 0.902 (0.897-0.937) &  10 &  10 \\ 
  Imagenette & PAWS & 100 & best kmediods (t-SNE) & 0.939 (0.931-0.942) &  10 &  10 \\ 
  Imagenette & PAWS & 100 & random & 0.917 (0.900-0.937) &  10 &  10 \\ 
  Imagenette & PAWS & 100 & random (class balanced) & 0.938 (0.908-0.940) &  10 &  10 \\ 
  Imagenette & supervised &  &  & 0.918 &   1 &  \\ 
   \hline
\end{tabular}

\normalsize
\end{table}

\FloatBarrier
\pagebreak

\bibliography{neurips_2023}

\end{document}